\documentclass[1p, preprint]{elsarticle}

\usepackage{amsmath}%
\usepackage{amsthm}
\usepackage{amsfonts}%
\usepackage{amssymb}%
\usepackage{mathrsfs}
\usepackage[capitalize]{cleveref}
\usepackage{overpic}
\usepackage{algpseudocode,algorithm}
\usepackage{url}
\newtheorem{theorem}{Theorem}
\newtheorem{lemma}{Lemma}
\newtheorem{proposition}{Proposition}
\newcommand{\R}{\mathbb{R}}

\renewcommand{\L}{\mathcal{L}}

\usepackage[utf8]{inputenc}
\usepackage{mathtools}
\usepackage{enumitem}
\usepackage{cprotect}

\usepackage{lineno}
\usepackage{bm, booktabs, multirow}

\DeclareMathOperator*{\argmax}{arg\,max}
\newcommand{\tr}[1]{\operatorname{tr} \left( #1 \right)}
\usepackage{xcolor}
\newcommand{\edit}[1]{{\color{black} #1}} % Edit
\newtheorem{innercustomthm}{Theorem}
\newenvironment{customthm}[1]
  {\renewcommand\theinnercustomthm{#1}\innercustomthm}
  {\endinnercustomthm}

\newtheorem{innercustomprop}{Proposition}
\newenvironment{customprop}[1]
  {\renewcommand\theinnercustomprop{#1}\innercustomprop}
  {\endinnercustomprop}

\newtheorem{innercustomlem}{Lemma}
\newenvironment{customlem}[1]
  {\renewcommand\theinnercustomlem{#1}\innercustomlem}
  {\endinnercustomlem}

\journal{Computer Physics Communications}

\begin{document}

\begin{frontmatter}

\cprotect\title{\verb|chebgreen|: Learning and Interpolating Continuous Empirical Green's Functions from Data}

\author[CEE]{Harshwardhan Praveen\corref{mycorrespondingauthor}}
\cortext[mycorrespondingauthor]{Corresponding author}
\ead{hp477@cornell.edu}

\author[CAM]{Jacob Brown}
\ead{jtb257@cornell.edu}

\author[CEE,CAM]{Christopher Earls}
\ead{earls@cornell.edu}

\address[CEE]{School of Civil and Environmental Engineering, Cornell University, Ithaca, NY, 14853, USA}
\address[CAM]{Center for Applied Mathematics, Cornell University, Ithaca, NY, 14853, USA}

\begin{abstract}
In this work, we present a mesh-independent, data-driven library, \verb|chebgreen|, to mathematically model one-dimensional systems, possessing an associated control parameter, and whose governing partial differential equation is \textit{unknown}. The proposed method learns an Empirical Green's Function for the associated, but hidden, boundary value problem, in the form of a Rational Neural Network from which we subsequently construct a bivariate representation in a Chebyshev basis. We uncover the Green's function, at an unseen control parameter value, by interpolating the left and right singular functions within a suitable library, expressed as points on a manifold of Quasimatrices, while the associated singular values are interpolated with Lagrange polynomials. This work improves upon prior work by extending the scope of applicability to non-self-adjoint operators and improves data efficiency.
\end{abstract}

\begin{keyword}
Green's function, PDE learning, Chebyshev Polynomials, Singular Value Expansion, Manifold interpolation
\end{keyword}

\end{frontmatter}

\section{Introduction}

Partial differential equations (PDEs) have always had a pivotal role in succinctly describing natural phenomena. They allow us to mathematically express the governing fundamental laws, as rate forms \cite{Feynman1967_Physical}. Researchers have worked on data-driven approaches and deep learning techniques to solve PDEs for various systems \cite{Raissi2018_Deep, Raissi2018_Hidden, Berg2017_Neural, Lu2021_DeepXDE, Sirignano2018_DGM, Weinan2018_DeepRitz}. Although they hold significant practical importance, the exact PDEs governing many important phenomena remain unknown across fields, including physics, chemistry, biology, and materials science. Thus, there also has been significant work in discovering \cite{Raissi2019_Physics, Schaeffer2017_learning, Brunton2016_Discovering, Champion2019_Discovery, Rudy2017_Data, Bonneville2022_Bayesian, Stephany2022_PDE_READ, Stephany2024_PDE_LEARN, Stephany2024_Weak_PDE_LEARN} the underlying PDEs from observational data. \textit{Operator Learning} represents another significant research avenue; aiming to approximate the solution operator associated with a hidden partial differential equation \cite{Boulle2024_Operator}. Notable contributions in this field include, but are not limited to, Deep Operator Networks \cite{Lu2021_DeepONet} which learns a branch (input encoding) and trunk (output encoding) networks to efficiently and accurately learn nonlinear operators from data, Fourier Neural Operators \cite{Li2020_FNO} which efficiently learn operators by parameterizing the integral kernel directly in Fourier space, Graph Neural Operator \cite{Li2020_GraphNO} which focus on capturing the Green kernel's short-range interactions to reduce the computational complexity of integration by using message passing on Graph Neural Networks, Multipole Graph Neural Operator \cite{Li2020_MGNO} which enhance graph-based methods by decomposing the Green kernel into a sum of low-rank kernels to capture long-range dependencies, and Geometry-informed Neural Operator \cite{Li2023_GINO}, which incorporate geometric information into the operator learning framework. A detailed overview of operator learning can be found in this survey article \cite{Boulle2024_Operator}.

In this paper, our focus is on approximating the \textit{Green's function} corresponding to some unknown governing differential equations for a given system of practical interest \cite{Evans2010_PDE}. The Hilbert-Schmidt Theorem \cite{Khanfer2024} provide a powerful framework for analyzing how Green’s functions associated with hidden differential operators, may be approximated. Within this setting, the Singular Value Expansion provides a valuable method for studying this approximation, particularly in the context of solving inverse problems and learning from data \cite{Fasshauer2015_Kernel}. Our proposed approach to operator learning is of practical value,
as it allows for the construction of data-driven models that inherently respect underlying physical laws governing a system of interest; effectively solving differential equations in a mesh-free manner. There has been recent theoretical work on elucidating learning rates, \emph{etc.} for Green's functions learned from data \cite{Boulle2021_Theory, Boulle2021_SVD, Boulle2022_ParabolicPDE}, as pertains to systems governed by elliptic or parabolic PDEs. Researchers have also utilized Deep Neural Networks to learn the Green's function in cases where the underlying operators are weakly non-linear. This approach involves employing a dual auto-encoder architecture to uncover latent spaces where lifting the response data somewhat linearizes the problem, thereby making it more tractable for Green's function discovery \cite{Gin2021_DeepGreen}. The field has also advanced through the development of Rational Neural Networks \cite{Boulle2020_Rational}, which have been applied to uncover Green's functions, offering mechanistic insights into the underlying system \cite{Boulle2021_Greenlearning}.

The work on learning \textit{Empirical Green's Functions} \cite{Praveen2023_EGF} introduces a method for using observational data to learn Green's functions, in self-adjoint contexts, without relying on machine learning techniques. The approach described in that work proposes learning \textit{Empirical Green's function} (EGF) from data in a discrete form,
    \begin{equation}
        \begin{aligned}
            \bm{G} &= \bm{U} \bm{\Sigma} \bm{U}^\top,\\
            \bm{G} \in \mathbb{R}^{N_{\text{sensors}}\times N_{\text{sensors}}} &\text{, } \bm{U} \in \mathbb{R}^{N_{\text{sensors}}\times K}, \bm{\Sigma} \in \mathbb{R}^{K \times K}
        \end{aligned}
    \end{equation}

That same work also proposes a methodology for interpolating these discrete EGF in a principled manner using a manifold interpolation scheme, which derives from the interpolating method for Reduced-order models \cite{Amsallem2008_ROM}.

In the present work, we utilize the deep learning library, \verb|greenlearning|, to learn Green's functions from data \cite{Boulle2021_Greenlearning}, and \verb|chebfun|, the open-source package for computing with functions to high accuracy, in order to improve the work on interpolating \textit{Empirical Green's Functions} \cite{Praveen2023_EGF} for 1D linear operators by:
\begin{itemize}
    \item Extending to non-self adjoint solution operator contexts.
    \item Reducing the data requirements to learn a Green's function.
    \item Introducing a completely mesh-independent formulation for representing the Green's function.
    \item Extending the interpolation method to interpolate the continuous representation of the Green's function.
\end{itemize}

With \verb|greenlearning|, we learn the green's function for a given problem using a synthetic dataset which consists of pairs of forcing functions and the system responses corresponding to those forcing functions. We learn a Neural network, $\mathcal{N}(x,y)$, which approximates the Green's function $G(x,y)$, $\forall x,y \in \Omega \subset \mathbb{R}$ . We propose creating a low-rank approximation for the neural network, $\mathcal{N}(x,y)$, using our \verb|Python| implementation of \verb|chebfun2| (\verb|chebfun| in two dimensions \cite{Townsend_Chebfun2}), which approximates bivariate functions using Chebyshev polynomials. \verb|chebfun2| allows us to construct a \textit{Singular Value Expansion} (SVE) for the learned Neural network approximating the Green's function in the form:

    \begin{equation}
        \mathcal{G}(x,y) = \mathcal{U}_{\infty \times K} \Sigma_{K \times K} \mathcal{V}_{\infty \times K}^*
    \end{equation}

\begin{figure}[htbp]
\centering
\begin{overpic}[width=\textwidth]{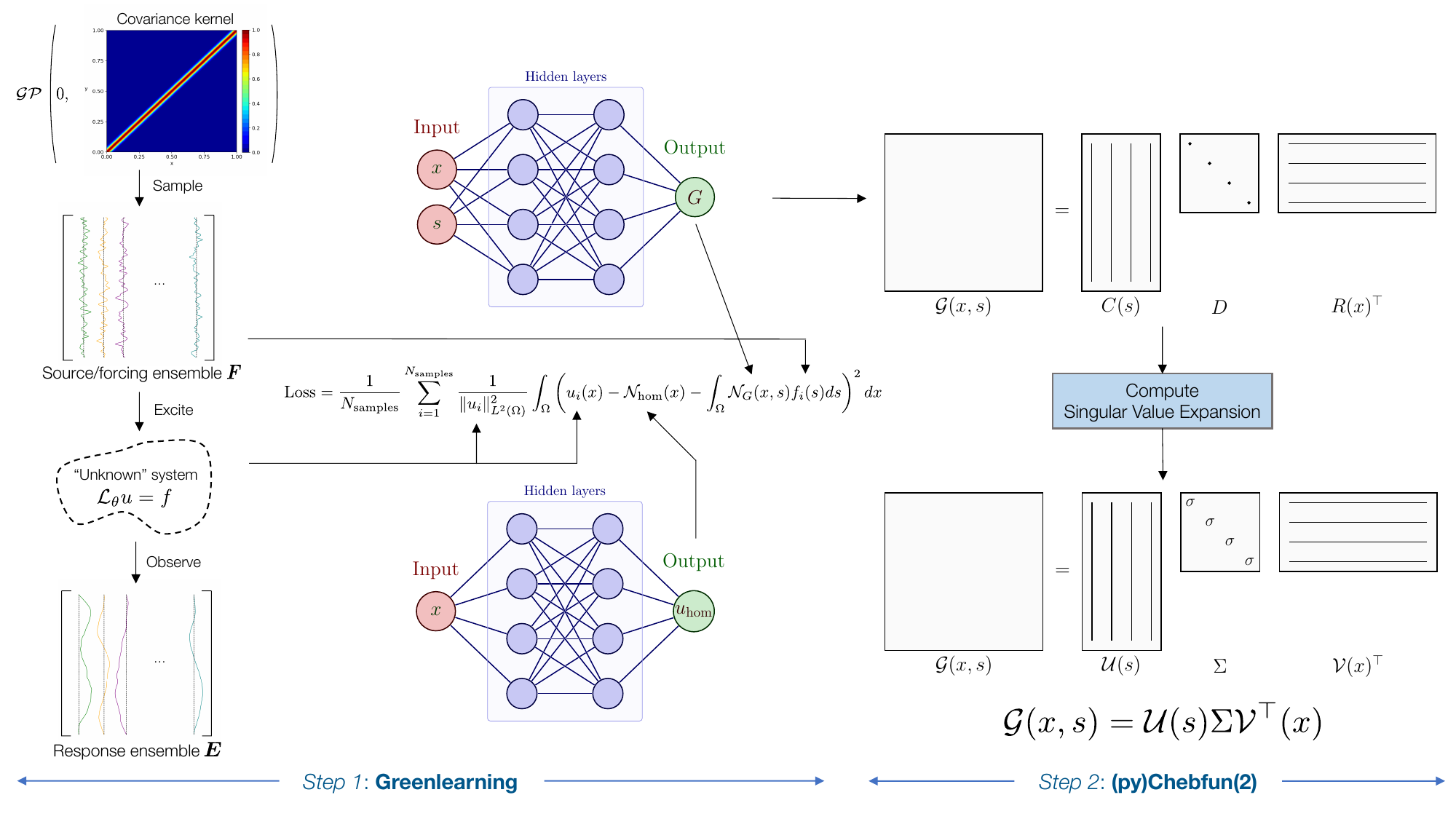}
\end{overpic}
\cprotect\caption{Schematic of the method for approximating Green's functions with \verb|chebgreen|: First, we approximate the Green's function from data using \verb|greenlearning| \cite{Boulle2021_Greenlearning}; In the second step, our \verb|Python| implementation of \verb|chebfun| is used to compute a Singular Value Expansion for the learned Green's function.}
\label{fig:method-schematic}
\end{figure}

Note that $\mathcal{U}$ and $\mathcal{V}$ are Quasimatrices (a ``matrix" in which one of the dimensions is discrete but the other is continuous), and $(\quad)^*$ denotes the adjoint of the Quasimatrix. Using this SVE, we interpolate the learned Green's function using an analogue of our interpolation algorithm from the previous work \cite{Praveen2023_EGF}. This representation, combined with the proposed algorithm, enables interpolation of the Green's function without discretizing the space, a requirement in previous work \cite{Praveen2023_EGF}. Additionally, it allows storing the Green's function in a form suitable for high-precision computing \cite{Chebfun}. All of this is implemented as a \verb|Python| package, called \verb|chebgreen|. \cref{fig:method-schematic} provides a brief overview of the proposed method.

In \cref{sec:methodology}, we provide a brief introduction to Green's functions, outline the process of generating our datasets, and present the proposed methods for learning and interpolating Green's functions. \cref{sec:numerical-examples} features various numerical benchmarks for the method in one dimension, both in the presence and absence of noise. Lastly, concluding remarks are presented in \cref{sec:conclusion}.

\section{Methodology} \label{sec:methodology}

Consider a linear differential operator, $\mathcal{L}_{\bm{\theta}}$, specified with a set $\bm{\theta} = (\theta_1,\ldots,\theta_{N_{\text{params}}}) \in \R^{N_{\text{params}}}$ of modeling parameters, and defined on a bounded and connected domain $\Omega\subset \R$ in one dimension, assumed to govern a physical system in the form of a boundary value problem:
    \begin{equation} \label{problem}
        \begin{aligned}
        \mathcal{L}_{\bm{\theta}} u &= f, &&\text{on } \Omega, \\
        \mathcal{B} u &= g, &&\text{in } \partial \Omega, 
        \end{aligned}
    \end{equation}
where, $\mathcal{B}$ is a linear differential operator specifying the boundary conditions of the problem ($g$ being the constraint on the problem boundary, $\partial \Omega$), $f$ is a forcing (or source) term, and $u$ is the unknown system response. It is pointed out that, while only a single boundary condition is shown in \cref{problem}, in an actual problem setting there would be a sufficient number of such conditions to ensure well-posedness. Under suitable conditions on the operator, there exists a Green's function $G_{\bm{\theta}}:\Omega\times\Omega\to\R$ associated with $\mathcal{L}_{\bm{\theta}}$ that is the impulse response of the linear differential operator and defined \cite{Evans2010_PDE} as:~
    \begin{equation*}
        \mathcal{L}_{\bm{\theta}} G_{\bm{\theta}}(x,s) = \delta (s - x), \quad x,s \in \Omega,
    \end{equation*}
where, $\mathcal{L}$ acts on the first variable and $\delta$ is the Dirac delta function. Then with homogeneous Dirichlet boundary conditions, \emph{i.e.}, $\mathcal{B} u = 0$ on the boundary of the domain, the solution to \cref{problem}, for a forcing, $f$, can be expressed using the Green's function within a \textit{Fredholm integral equation} of the first-kind:

    \begin{equation*}
        \Tilde{u}(x) = \int_\Omega G_{\bm{\theta}}(x,s) f(s) d s, \qquad x,s \in \Omega.
    \end{equation*}

Now, if we consider the homogeneous solution to \cref{problem}, $u_{\text{hom}}$, which can be found by solving the following boundary value problem:~
    \begin{align*}
        \mathcal{L}_{\bm{\theta}} u_{\text{hom}} &= 0, \\
        \mathcal{B} u &= g.
    \end{align*}
we can use superposition, to construct solutions, $u$, to \cref{problem} as $u = \Tilde{u} + u_{\text{hom}}$,
    \begin{equation}
        u(x) = \int_\Omega G_{\bm{\theta}}(x,s) f(s) d s + u_{\text{hom}}, \qquad x,s \in \Omega
    \end{equation}

In this work, we only consider examples where $\bm{\theta} = \theta \in \mathbb{R}$, or $N_{\text{param}} = 1$, but all the proofs and methods can be trivially extended to work for $N_{\text{param}} \ge 2$. We begin with a description of the data (in our case synthetic data generated from simulations) which we use to learn the Green's functions.

\subsection{Generating the dataset} \label{sec:dataset-generation}
The datasets used to learn Green's functions consist of $N_{\text{samples}}\geq 1$ forcing terms, $\{f_j\}_{j=1}^{N_{\text{samples}}}$, and the corresponding system's responses, $\{u_j\}_{j=1}^{N_{\text{samples}}}$, satisfying \cref{problem}. In the present work, our systems~\eqref{problem} are forced with random functions, sampled from a Gaussian process (GP), $\mathcal{GP}(0,K)$, with a covariance kernel, $K$, as motivated by recent theoretical results \cite{Boulle2021_Theory}. More specifically, the forcing terms are drawn from a GP, with mean zero and squared-exponential covariance kernel, $K_{\text{SE}}$, defined as:~
    \begin{equation*}
        K_{\text{SE}}(x, s) = \exp \left(-\dfrac{|x-s|^2}{2 l^2} \right), \quad x, s \in \Omega,
    \end{equation*}
where, $\Omega = [a,b] \subset\R$ is the one-dimensional problem domain and $l>0$ is the length scale hyperparameter that sets the correlation length for this kernel. Similar to \verb|greenlearning| \cite{Boulle2021_Greenlearning}, we define a normalized length scale parameter, $\sigma = l/(b - a)$, in order to remove the dependence on the length of the domain. This parameter, $\sigma$ is chosen to be larger than the spatial discretization at which we are numerically evaluating the forcing terms; to ensure that the random functions are resolved properly within the discrete representation employed. Additionally, $\sigma$ is specified to possess a suitable magnitude in order to ensure that the set of forcing terms is of full numerical rank. Note that other types of covariance kernels used in recent deep learning studies~\cite{Li2021_Fourier}, such as Green's functions related to Helmholtz equations, are likely to produce improved approximation results, as they inherently contain some information about the singular vectors of the operator, $\L_{\theta}$. This is reflected in the theoretical bounds for the randomized SVD with arbitrary covariance kernels, which show that one may obtain higher accuracy by incorporating knowledge of the leading singular vectors of the differential operator into the covariance kernel~\cite{Boulle2021_SVD}. In the present work, we employ a generic squared-exponential kernel as one may not have prior information about the governing operator, $\L_{\theta}$, in real applications. In the case of problems with a periodic boundary conditions we draw the forcing functions from the periodic kernel:

\begin{equation*}
    K_{\text{Periodic}}(x, s) = \exp \left(-\dfrac{2\sin^2(\pi |x-s| / p)}{l^2} \right), \quad x, s \in \Omega,
\end{equation*}
where $p$ is the desired period of the functions.

We are interested in applications where one can only measure the responses at a finite number of locations, $\bm{x} = \{x_1, \ldots, x_{N_{x}}\}$, where ${x}_i \in \Omega$ and $N_{x}\geq 1$ is the number of measurements taken within the domain, $\Omega$. Similarly, we sample the forcing terms at $\bm{s} = \{s_1, \ldots, s_{N_{s}}\}$, where ${s}_i \in \Omega$ and $N_{s}\geq 1$ to assemble the following column vectors:
    \begin{equation*}
        f_i = f_i(\bm{s}) \coloneqq \begin{bmatrix} f_i(s_1) & \cdots & f_i(s_{N_{s}})\end{bmatrix}^\top, \quad
    u_i = u_i(\bm{x};\theta) \coloneqq \begin{bmatrix} u_i(x_1) & \cdots & u_i(x_{N_{x}})\end{bmatrix}^\top,
    \end{equation*}
The notation $u_i(\bm{x};\theta)$ highlights the dependence of the responses on the parameter $\theta$, and $\top$ denotes the matrix transpose. We collect these as $N_{\text{samples}}$ number of input-output pairs as $\{f_i, u_i\}$. In order to gauge fidelity of the learned Green's function to the true underlying Green's function, during training, we leave out a part of our dataset ($\sim 5\%$) to serve as a validation dataset and don't use this during our learning process.

\begin{equation*}
    \{f_i, u_i\}_{i = 1}^{N_{\text{samples}}} = \{f_i, u_i\}_{i = 1}^{N_{\text{train}}} + \{f_i, u_i\}_{i = 1}^{N_{\text{validation}}}
\end{equation*}

\subsection{Low-rank approximation of the Green's function}

We want to compute a low-rank approximation for the Green's function $G_\theta$ associated with the linear differential operator, $\mathcal{L}_\theta$, by employing a singular value expansion (SVE) :
    \begin{equation}
        \mathcal{G}_\theta(x,y) = \sum_{k = 1}^{K} \sigma_k \phi_k (y) \psi_k (x),
    \end{equation}
where, $\sigma_1 \ge \dots \ge \sigma_K \ge 0$ are the singular values, $\{\phi_k\}_{k = 1}^{K}$ are the left singular functions, and $\{\psi_k\}_{k = 1}^{K}$ are the right singular functions. Here, each term $\sigma_k \phi_k (y) \psi_k (x)$ is a \textit{rank} 1 approximation which can be thought of as an ``outer product" of two univariate functions.

We choose a truncated Singular Value Expansion (SVE) because the SVE of an infinite-dimensional Hilbert-Schmidt (HS) operator so long as the Green's function is square-integrable), after retaining $K$ terms, yields the best rank-$K$ approximation under the HS and operator norms; by the Eckart-Young-Mirsky theorem \cite{Stewart1990_SVE}. This representation is important because it allows us to interpolate learned Green's functions between different parametric values (\cref{subsec:interp}). Moreover, in practice, we learn a rational approximation to the Green's function as described in the next section, which has exponentially decaying singular values. This makes an SVE the appropriate choice for representing a learned Green's functions.

\subsection{Approximating Green's functions with Neural Networks}

In order to compute an efficient approximation for the Green's function, $\mathcal{G}_\theta (x,y)$, we use the method developed by Boull\'e et al. \cite{Boulle2021_Greenlearning}. The main idea is to train two Rational Neural Networks \cite{Boulle2020_Rational}: $\mathcal{N}_{\text{G}} (x,s)$ which approximates the Green's function associated with the underlying linear differential, $\mathcal{L}_\theta$; and $\mathcal{N}_{\text{hom}} (x)$ which corresponds to the homogeneous solution associated with the boundary conditions, by minimizing the following loss over our training dataset $\{f_i, u_i\}_{i = 1}^{N_{\text{train}}}$:

\begin{equation}
    \text{Loss} = \frac{1}{N_{\text{train}}} \sum_{i = 1}^{N_{\text{train}}} \frac{1}{\|u_i\|_{L^2(\Omega)}^2} \int_\Omega \left( u_i (x) - \mathcal{N}_{\text{hom}} (x) - \int_\Omega \mathcal{N}_{\text{G}} (x,s) f_i (s) d s \right)^2 d x
\end{equation}

Rational Neural Networks offer a significant advantage for learning Green’s functions: their poles offer a natural means for approximating singularities within underlying Green’s functions and corresponding homogeneous solutions \cite{Boulle2021_Greenlearning}. Additionally, their superior approximation properties, validated both theoretically and empirically \cite{Boulle2020_Rational}, make them particularly well-suited for this task.

We propose a slight modification to the loss function for problems where we know that the boundary conditions are Dirichlet. In these case, the Green's function is zero on the boundaries. More explicitly, if the problem is defined on the domain $\Omega = [a,b] \subset \mathbb{R}$, and the boundary conditions are Dirichlet,

\begin{equation*}
    G_\theta(a,s) = G_\theta(b,s) = G_\theta(x,a) = G_\theta(x,b) = 0, \quad x,s \in \Omega.
\end{equation*}

Incorporating this information in the loss function improves the fidelity of the computed left and right singular functions (described in \cref{sec-singular-value-expansion}) near the boundaries of the domain. We utilize \textit{Approximate Distance Functions} \cite{Sukumar2022_Distancefunctions} to enforce this by changing the loss function slightly to

\begin{equation} \label{eq:loss-function-greenlearning}
    \text{Loss} = \frac{1}{N_{\text{train}}} \sum_{i = 1}^{N_{\text{train}}} \frac{1}{\|u_i\|_{L^2(\Omega)}^2} \int_\Omega \left( u_i (x) - \mathcal{N}_{\text{hom}} (x) - \int_\Omega \alpha (x,s) \mathcal{N}_{\text{G}} (x,s) f_i (s) d s \right)^2 d x,
\end{equation}
where, $\alpha(x,s)$ is the Approximate Distance Function for a rectangular domain. To construct this, we start from the approximate distance function for a line segment \cite{Rvachev2001_LineADF} that joins $\bm{x}_1 \equiv (x_1, s_1)$ and $\bm{x}_2 \equiv (x_2, s_2)$, with a center $\bm{x}_c \equiv (\bm{x}_1 + \bm{x}_2)/2$ and length $L = \| \bm{x}_1 - \bm{x}_2\|$:

\begin{align*}
    \beta \equiv \beta(\bm{x}) = \sqrt{h^2 + \left( \dfrac{\varphi - t}{2} \right)^2}, \quad  \varphi = \sqrt{t^2 + h^4},
\end{align*}
where,
\begin{align*}
    h \equiv h(\bm{x}) &= \dfrac{(x-x_1)(s_2 - s_1) - (s-s_1)(x_2-x_1)}{L}, \\
    t \equiv t(\bm{x}) &= \dfrac{1}{L} \left[ \left(\dfrac{L}{2}\right)^2 - \|\bm{x}-\bm{x}_c\|^2\right].
\end{align*}

\begin{figure}[htbp]
\centering
\begin{overpic}[width=0.7\textwidth]{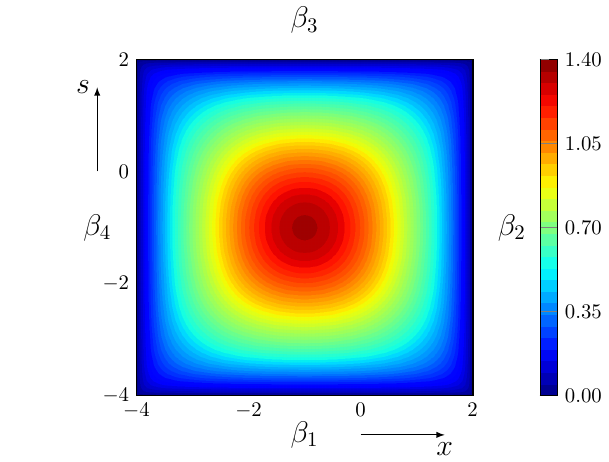}
\end{overpic}
\cprotect\caption{Example of the approximate distance function for a domain $[-4,2] \times [-4,2]$.}
\label{fig:aproximate-distance-function}
\end{figure}

We then combine the approximate distance functions for the four segments, $\{\beta_1, \beta_2, \beta_3, \beta_4 \}$ with an R-equivalence solution that preserves normalization up to order $m = 1$ of the distance function at all regular points, given by \cite{Biswas2004_Requiv}:

\begin{align*}
    \alpha(x,s) = \beta(\beta_1, \beta_2, \beta_3, \beta_4) = \dfrac{1}{\sqrt[m]{\dfrac{1}{\beta_1^m} + \dfrac{1}{\beta_2^m} + \dfrac{1}{\beta_3^m} + \dfrac{1}{\beta_4^m}}}
\end{align*}

An example of the approximate distance function is shown in \cref{fig:aproximate-distance-function}. In the cases where the problem does not have a Dirichlet boundary condition or the boundary conditions are unknown, we set $\alpha(x,s) \equiv 1$.
In this context, ``approximate" refers to distance functions approximating the exact signed distance function, which
are, themselves, the orthogonal distance from a point to the boundary of a set. The adopted approximation implicitly specifies a Dirichlet boundary condition. However, this is not a requirement; Rational Neural Networks can learn the boundary conditions directly, from training data \cite{Boulle2021_Greenlearning}. When Dirichlet boundary conditions are known a priori, employing approximate distance functions improves the numerical approximation of the Green’s functions near the boundaries, which in turn improves the numerical approximation of the learned singular functions (\cref{sec-singular-value-expansion}).

To compute the loss function described in \eqref{eq:loss-function-greenlearning}, the integrals are discretized using a trapezoidal rule \cite{Endre2003_NumericalAnalysis} over the measurement locations for the forcing terms $\{s_i \}_{i = 1}^{N_s}$ and the system responses $\{x_i \}_{i = 1}^{N_x}$. The trapezoidal rule is used for numerical integration because the numerical experiments are done on a uniform grid. However, a uniform grid is not a requirement; the underlying implementation allows users to specify arbitrary sensor locations for both forcing terms and system responses as detailed in \cref{sec:dataset-generation}. Alternatively, users can leverage the implementation Monte Carlo integration in the library if sensor locations are non-uniformly distributed. Prior work on learning Green's functions with Rational Neural Networks \cite{Boulle2021_Greenlearning} has demonstrated the robustness of Rational Neural Networks to variations in sensor placement and the choice of quadrature rule for integration. This gives us a bivariate function as a product of Rational Neural Network with a fixed function (which is defined purely in terms of the domain boundaries), $g(x,s) = \alpha(x,s) \mathcal{N}_{\text{G}} (x,s)$, which approximates the Green's function associated with our problem. Now we would like to construct a singular value expansion (SVE) for this bivariate function. To do so, we implemented the necessary parts of the \verb|chebfun| \verb|MATLAB| library \cite{Chebfun} in \verb|Python| as \verb|chebpy2|, using the 1D implementation, \verb|chebpy| \cite{Chebpy}. In the next two subsections, we describe the basic details of how \verb|chebfun| constructs a SVE for bivariate functions using polynomial interpolants in a Chebyshev basis.

\subsection{Approximating a bivariate function, $g(x,y)$, with \textit{chebfun2}} \label{subsec:chebfun2-aprox}

A \verb|chebfun| is a polynomial interpolant of a smooth function $f: [-1,1] \to \mathbb{R}$ evaluated at $n+1$ Chebyshev points, and furnished with:

    \begin{equation*}
        x_i = \cos \left( \frac{i \pi}{N} \right), \quad 0 \le i \le N
    \end{equation*}

After adaptively choosing a polynomial degree, $N$, to approximate $f$ to machine precision, a \verb|chebfun| stores these $N+1$ values $(f(x_i))_{0 \le i \le N}$ as a vector, whose entries are the coefficients for a Lagrange basis,

    \begin{equation*}
        f(x) \approx \sum_{i = 0}^{N} f(x_i) l_i(x), \quad l_i(x) = \frac{\prod_{j = 0, j \ne i}^{N} (x - x_j)}{\prod_{j = 0, j \ne i}^{N} (x_i - x_j)}
    \end{equation*}

Approximating a function with a \verb|chebfun| allows for constructing infinite-dimensional analogues of matrices, which are called \textit{Quasimatrices}. A \textit{column Quasimatrix} can be thought of as $\infty\times K$ matrices. In other words, they have finitely many columns, each column being represented by a square-integrable function, in this case a \verb|chebfun|.
    \begin{figure}[!htb]
        \centering
        \includegraphics[width=0.7\linewidth]{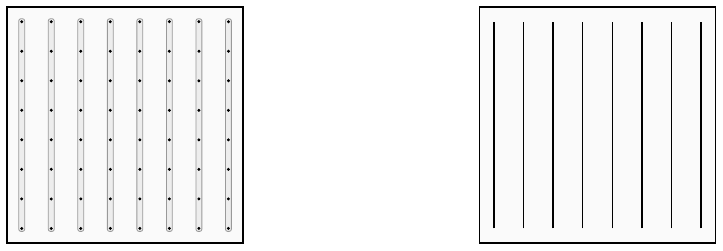}
        \cprotect\caption{Entries in the column, represented by dots, of a matrix (left) can be replaced by a \verb|chebfun|, represented by a solid line, in a Quasimatrix (right)}
        \label{fig:matrix-quasimatrix}
    \end{figure}

Consider a $\infty \times K$ column Quasimatrix $\mathcal{A}$, where $\{\mathcal{A}_k \}_{k = 1}^{K}$ are \verb|chebfun|-s on some domain $\Omega \subset \mathbb{R}$.
    \begin{equation}
            \mathcal{A}:=\left[
            \begin{array}{ccc}
            \mathcal{A}_1 & \dots & \mathcal{A}_K
            \end{array}\right]
    \end{equation}
We may also view this Quasimatrix $\mathcal{A}$ as an element of the vector space $(L^2(\Omega))^K$. This is a natural space to look for low-rank approximations to a Hilbert-Schmidt integral operator such as those associated with certain Green's functions, as they are compact operators on the Hilbert space $L^2(\Omega)$. We start by defining a few operations on the Quasimatrix. For some scalar $c$,
\begin{equation*}
    c \mathcal{A} = \mathcal{A} c:=\left[
            \begin{array}{ccc}
            c \mathcal{A}_1 & \dots &  c \mathcal{A}_K
            \end{array}\right]
\end{equation*}
For a $K \times K$ matrix $C$, we can define its product with a Quasimatrix as:
\begin{equation*}
    \mathcal{A} C:=\left[
            \begin{array}{ccccc}
            \sum_{j = 1}^{K} C_{j1}\mathcal{A}_j & \dots & \sum_{j = 1}^{K} C_{jk}\mathcal{A}_j & \dots & \sum_{j = 1}^{K} C_{jK}\mathcal{A}_j
            \end{array}\right]
\end{equation*}
Similarly, for a \textit{row Quasimatrix} $\mathcal{B} :=\left[\begin{array}{ccc}\mathcal{B}_1 & \dots & \mathcal{B}_K \end{array}\right]^\top $, which can be thought of as a $K \times \infty$ matrix, such that each row is a \verb|chebfun|, one can define:
\begin{equation*}
    C \mathcal{B}:=\left[
            \begin{array}{c}
            \sum_{j = 1}^{K} \mathcal{B}_j C_{1j} \\
            \vdots \\
            \sum_{j = 1}^{K} \mathcal{B}_j C_{kj} \\
            \dots \\
            \sum_{j = 1}^{K} \mathcal{B}_j C_{Kj}
            \end{array}\right]
\end{equation*}

In this paper, we will denote the adjoint of a column Quasimatrix $\mathcal{A}$ as $\mathcal{A}^*$, and note that, given another column Quasimatrix $\mathcal{C}$, \textit{the inner product} between two quasimatrices (an outer product on the space $(L^2(\Omega))^K$) is given by \cite{Altmann2022_Stiefel}:
    \begin{equation}
            \mathcal{A}^*\mathcal{C}:=\left[
            \begin{array}{ccc}
            \left(\mathcal{A}_1, \mathcal{C}_1\right)_{L^2(\Omega)} & \cdots & \left(\mathcal{A}_1, \mathcal{C}_K\right)_{L^2(\Omega)} \\
            \vdots & \ddots & \vdots \\
            \left(\mathcal{A}_K, \mathcal{C}_1\right)_{L^2(\Omega)} & \cdots & \left(\mathcal{A}_K, \mathcal{C}_K\right)_{L^2(\Omega)}
            \end{array}\right] \in \mathbb{R}^{K \times K}
    \end{equation}
Although we are not considering the case of Quasimatrices with complex-valued functions as columns, we use the notation $\mathcal{A}^*$ rather than $\mathcal{A}^\top$ in order to distinguish from the usual matrix transpose, as both of these operations show up in the proofs in the Appendix.

The idea of employing a polynomial approximation of a univariate function on a Chebyshev grid can be extended to approximate bivariate functions. \verb|chebfun2| \cite{Townsend_Chebfun2} constructs approximate $g(x,y) \approx C(y) D R(x)$\footnote{This is analogous to a CUR-decomposition for a matrix but we use $D$ instead of $U$ to avoid confusion with our terminology for left singular vectors}. It constructs a rank $K$ approximant as follows:

    \begin{equation} \label{eq: chebfun2-approx}
        g(x,y) \approx g_K (x,y) = \sum_{k = 1}^{K} D_k C_k (y) R_k (x)
    \end{equation}

Here, each $C_k,R_k$ is a \verb|chebfun| and $D_k \in \mathbb{R}$ is a pivot value. The $Chebfun$-s $C_k$ are collected into one quasimatrix $C(y) = [C_1(y), C_2(y), \dots, C_K(y)]$, and $R_k$ into another quasimatrix $R(x) = [R_1(x), R_2(x), \dots, R_K(x)]$.

While constructing the approximation, \verb|chebfun2| chooses pivot, $(x_k,y_k)$ on the domain of the bivariate function and constructs a column \verb|chebfun| $C_k(y)$ to approximate $G(x_k, y)$ and a row \verb|chebfun| $R_k(x)$ to approximate $G(x, y_k)$.

    \begin{figure}[!htb]
        \centering
        \includegraphics[width=0.5\linewidth]{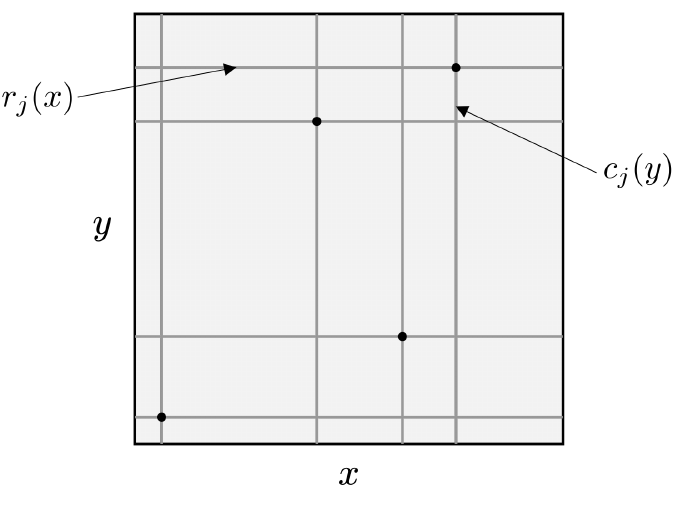}
        \caption{Example of pivot locations on a grid}
        \label{fig:pivot-locations}
    \end{figure}

The pivots are chosen at the locations where the absolute error between the current approximant and the bivariate function $G(x,y)$ is maximum. Repeating this procedure iteratively for $K$ steps, \verb|chebfun2| constructs an approximation for $g(x,y)$. A more detailed overview of the workings of \verb|chebfun2| can be found in the original work \cite{Townsend_Chebfun2}.

\subsection{Constructing a singular value expansion of $g(x,y)$} \label{sec-singular-value-expansion}

Note that the end-goal of using \verb|chebfun2| is to construct a singular value expansion for $G(x,y)$ as follow:

    \begin{equation*}
        g(x,y) = \mathcal{U}_{\infty \times K} (y) \Sigma_{K \times K} \mathcal{V}_{\infty \times K}^* (x)
    \end{equation*}

Constructing this Singular Value Expansion thus requires computing a QR decomposition of a Quasimatrix. The details of the same can be found in Trefethen's work on Householder triangularization of a Quasimatrix \cite{Trefethen2010_QR}. The main details are summarized here.

\begin{figure}[!htb]
        \centering
        \includegraphics[width=\linewidth]{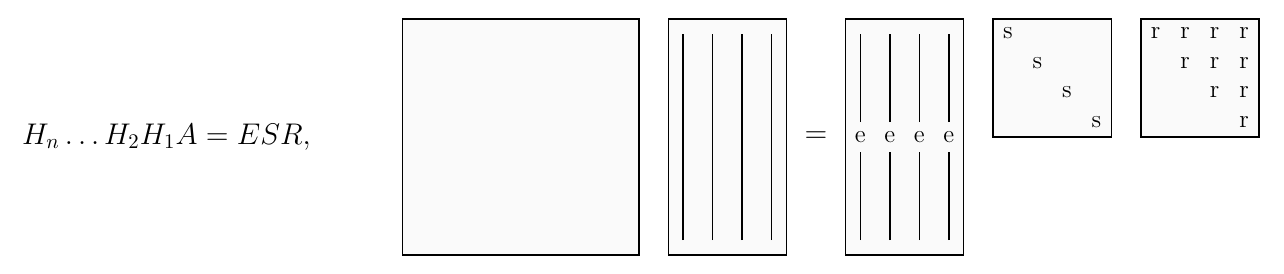}
        \caption{Householder triangularization of a Quasimatrix}
        \label{fig:QR-Quasimatrix}
    \end{figure}

Let's assume we want to find a QR decomposition for a $\infty \times n$ Quasimatrix $\bm{A}$, the columns of which are functions of $x \in [a,b]$ ($\bm{A}$ is a linear map from $\mathbb{C}^n$ to $L^2[a,b]$). Let the notation $\bm{v}^* \bm{w}$ denote the inner product and $\| \cdot \|$ be the $L^2$ norm on $[a,b]$. We would like to decompose this Quasimatrix as $\bm{A} = \bm{Q} \bm{R}$, such that $\bm{Q}$ is an orthonormal Quasimatrix and $\bm{R}$ is an upper triangular matrix. The aim is to compute this by the numerically more stable method of Householder reflections. This is done by applying a self-adjoint operator, $\bm{H}$, so that one can get the form as shown in \cref{fig:QR-Quasimatrix}. Here, the vertical lines denote that the column is a \verb|chebfun|.

$\bm{E}$ will be an $\infty \times n$ quasimatrix , fixed in advanced, with orthonormal columns $\{e_j\}$, and $\bm{S}$ is an $n \times n$ sign matrix, so as to ensure the diagonal of $\bm{R}$ is real and non-negative. In practice, $e_j$ is taken to be a multiple of the $(j-1)$th Legendre polynomial $P_{j-1}(x)$, scaled to $[a,b]$. The Householder reflector $\bm{H}_k$ is a self-adjoint operator acting on $L^2[a,b]$, chosen so as to map a certain function to another function of equal norm in the space spanned by $e_1, \dots, e_k$.  

The precise formulas to compute these components corresponding for each column $k \in 1, \dots, n$ are as follows:
    \begin{align*}
        &\bm{x} = \bm{A}[:,k], \quad \bm{R}_{kk} = \|\bm{x}\|, \quad \bm{S}_k = -\text{sign}(e^*_k \bm{x}), \\
        &\bm{y} = \bm{S}_k \bm{R}_{kk} e_k, \quad \bm{v} = \frac{\bm{y} - \bm{x}}{\|\bm{y} - \bm{x}\|},\quad \bm{H}_k = \bm{I} - 2 \bm{v} \bm{v}^*,\\
        &\bm{A}[:,j] = \bm{H}_k \bm{A}[:,j],\quad \bm{R}_{kj} = e^*_k \bm{A}[:,j],\quad \bm{A}[:,j] = \bm{A}[:,j] - \bm{R}_{kj} e_k,\quad k+1 \le j \le n.
    \end{align*}
Here, the outer product notation $\bm{v} \bm{v}^*$ denotes the operator that maps a function $\bm{w}$ to $\bm{v} (\bm{v}^* \bm{w})$. Since $\bm{H}_1 \cdots \bm{H}_n$ is the adjoint of $\bm{H}$ in $L^2[a,b]$, $\bm{Q} = \bm{H}_1 \cdots \bm{H}_n \bm{E} \bm{S}$. This gives us our QR decomposition.

The process of constructing an SVE with \verb|chebfun2| \cite{Townsend_Chebfun2} is outlined in \cref{sve-chebfun2}.

    \begin{algorithm}[htbp]
        \cprotect\caption{SVE using \verb|chebfun2|}
        \label{sve-chebfun2}
        \begin{algorithmic}
            \State \textsl{\textbf{Input:}} A \verb|chebfun| $G(x,y) \approx C(y) D R(x)$
            \State \textsl{\textbf{Output:}} Quasimatrices with orthonormal columns $\mathcal{U}_G$ and $\mathcal{V}_G$, and a diagonal matrix $\Sigma_G$
            \State 1. Householder triangularization: $C(y) = Q_C(y) R_C$, $R(x)^* = Q_R(x) R_R$.
            \State 2. Compute: $A = R_C D R_R^*$.
            \State 3. Matrix SVD: $A = U_A S_A V_A^*$.
            \State 4. Compute: $\mathcal{U}_G = Q_C(y) U_A$, $\mathcal{V}_G = Q_R(x) V_A$, and $\Sigma_G = S_A$.
        \end{algorithmic}
    \end{algorithm}
    
This gives us the singular value expansion we desire:
\begin{equation*}
    g(x,y) \approx \mathcal{U}_g (y) \Sigma_g \mathcal{V}^*_g (x)
\end{equation*}

\cprotect\subsection{\verb|chebgreen|: Combining \verb|greenlearning| and \verb|chebfun|} \label{sec:chebgreen-construction}
With these two key components, we can efficiently learn and store a singular value expansion for a Green's function, starting from data in the form of input (forcing functions) - output (system responses) pairs, for the ``unknown" underlying linear differential equation:

\begin{itemize}
    \item Learn two Rational Neural networks, $\mathcal{N}_{\text{G}} (x,s)$ and $\mathcal{N}_{\text{hom}} (x)$, from the training data, $\{f_i, u_i\}_{i = 1}^{N_{\text{train}}}$ to approximate the Green's functions as $g(x,y) = \alpha(x,s) \mathcal{N}_{G}(x,s)$ along with the  homogeneous solution, $u_{\text{hom}} = \mathcal{N}_{\text{hom}} (x)$.
    \item Use our \verb|Python| implementation of the \verb|chebfun| library to represent these approximations in a Chebyshev basis:
    \begin{itemize}
        \item We store the Green's function as a singular value expansion of a \verb|chebfun2|:
            \begin{align*}
                g(x,s; \theta) &= \alpha(x,s) \mathcal{N}_{G}(x,s; \theta), \\
                \mathcal{G}_{\theta}(x,s) &\equiv \mathcal{U} (y; \theta) \Sigma(\theta) \mathcal{V}^*(x;\theta) = \mathcal{U}_g (y) \Sigma_g \mathcal{V}^*_g (x) = g(x,s; \theta).
            \end{align*}
        \item The homogeneous solution is stored as a \verb|chebfun|:
            \begin{equation*}
                \mathcal{H}(x) \equiv \mathcal{N}_{\text{hom}} (x).
            \end{equation*}
    \end{itemize}
\end{itemize}
\subsection{Interpolating the solution on a manifold} \label{subsec:interp}

The algorithm proposed by Praveen et al.\cite{Praveen2023_EGF} relies on the method of interpolating in a tangent space of the compact Stiefel manifold. The compact $k-$Stiefel manifold of $\mathbb{R}^m$ is defined to be the manifold of equivalence classes of $m\times k$ matrices with orthonormal columns, where $U$ and $U'$ are consiedered equivalent when their columns span the same subspace of $\mathbb{R}^m$ \cite{Absil2008_Manifolds}. In other words, we may think of it as
    \begin{equation}
        \begin{aligned}
            S_K(\mathbb{R}^m) = \lbrace{U \in \mathbb{R}^{m\times K} \colon U^TU = I_K }\rbrace.
        \end{aligned}
    \end{equation}
    
Their previous algorithm takes the SVDs of known Green's functions, $G_i=\Phi_i \Sigma_i \Phi_i ^T$, at parameters $\theta_i$, chooses an index $j$ that is closest to the unseen parameter, $\theta^*,$ and lets $\Phi_j = \hat{\Phi}$ be the reference basis (base point) for the interpolation. The other known orthonormal matrices, $\Phi_i$, represented as elements on the Stiefel manifold, are then mapped to the tangent space at $\hat{\Phi}$ via an approximation to the logarithmic map \cite{Sternfels2013_Interp}. These points in the tangent space are then interpolated via Lagrange polynomial interpolation, and the matrix, $\Phi^*$, for the unseen parameter, $\theta^*$, is obtained by mapping the interpolated point back to the Stiefel manifold from the tangent space using an exponential mapping \cite{Sternfels2013_Interp}.

We wish to generalize this method to the case where the orthonormal matrices $\Phi$ are instead rank $K$ operators  represented by a quasimatrix whose $K$ columns are $L^2(\Omega)$ functions, where $\Omega$ is a suitable domain. In other words, whereas in the finite dimensional problem we were interested in $m\times K$ real matrices, here we are interested in the space $H= (L^2(\Omega))^K,$ which may also be thought of as $\mathcal{L}(\mathbb{R}^k, L^2(\Omega)),$ the space of linear transformations from $\mathbb{R}^K$ to $L^2(\Omega).$

In order to generalize the manifold interpolation, we first need to find a space analogous to the finite-dimensional Stiefel manifold, and to identify its tangent space \cite{Altmann2022_Stiefel}. We write elements, $\Phi$, of $H$ in row-vector notation as $\Phi = (\Phi_1,...,\Phi_K),$ so that we can multiply quasimatrices (viewed here as vectors in $H$) on the right by a $K\times K$ matrix to get another vector in $H.$ The space $H$ admits an outer product \cite{Altmann2022_Stiefel}
    \begin{equation}
        \begin{aligned}
            \Phi^* \Psi:=\left[\begin{array}{ccc}
\left(\Phi_1, \Psi_1\right)_{L^2(\Omega)} & \cdots & \left(\Phi_1, \Psi_K\right)_{L^2(\Omega)} \\
\vdots & \ddots & \vdots \\
\left(\Phi_K, \Psi_1\right)_{L^2(\Omega)} & \cdots & \left(\Phi_K, \Psi_K\right)_{L^2(\Omega)}
\end{array}\right] \in \mathbb{R}^{K \times K}
        \end{aligned}
    \end{equation}
as well as an inner product
    \begin{equation}
        \begin{aligned} (\Phi, \Psi)_H:=\sum_{j=1}^K\left(\Phi_j, \Psi_j\right)_{L^2(\Omega)}=\operatorname{tr} (\Phi^* \Psi).
        \end{aligned}
    \end{equation}
Here, we use the suggestive notation $\Phi^*  \Psi$ for the outer product on $H$ to emphasize that this is both distinct from, and analogous with, the matrix transpose \cite{Harms2012_Manifolds}. We then define
    \begin{equation}
        \begin{aligned}
            S_K(H)=\lbrace{\Phi\in H\colon \Phi ^* \Phi=I_K }\rbrace.
        \end{aligned}
    \end{equation}
The inner product on $H$ induces a Riemannian metric, and turns $S_K(H)$ into a Riemannian submanifold of $H$ which we will call the \textit{Stiefel manifold}, and will denote from here on simply by $S_K$ \cite{Altmann2022_Stiefel}. It can be shown that its tangent space at a point $\Phi$ is 
    \begin{equation}
        \begin{aligned}
            T_\Phi S_K = \lbrace{\Psi\in H\colon \Psi^* \Phi = -\Phi^* \Psi \rbrace}
        \end{aligned}
    \end{equation}
which is analogous to the identification of the tangent space for the finite dimensional Stiefel manifold.

The next step towards extending the manifold interpolation algorithm to the Quasimatrix case is to find an orthogonal projection mapping from $H$ to $T_\Phi S_K,$ the tangent space of the Stiefel manifold at $\Phi.$ We established the following result, which we will prove in the appendix of this paper:

\begin{theorem}
    The map $P_\Phi\colon H \rightarrow T_\Phi S_K$ given by $\Psi\mapsto \Psi - \Phi\operatorname{sym}(\Phi^*\Psi)$, where $\operatorname{sym}(\Phi^* \Psi) \coloneq \frac{1}{2}(\Phi^* \Psi + \Psi^* \Phi)$, is the orthogonal projection onto the tangent space of the Stiefel manifold $S_K$ at $\Phi.$
\end{theorem}

This projection mapping is our infinite dimensional analogue to Step 4 Part 1 of the previous algorithm \cite{Praveen2023_EGF}.

In order to extend the third part of Step 4 of that same algorithm to the Quasimatrix generalization, we need to establish a valid approximation to the exponential map so that we can map interpolated points in the tangent space back to the Stiefel manifold. The key concept is that of a retraction on a manifold, which may be thought of as a first order approximation to the exponential map. In their previous algorithm Praveen et al.\cite{Praveen2023_EGF} use QR factorization as their approximation to the exponential map, as it is retraction on the finite dimensional Stiefel manifold. Altman et al. \cite{Altmann2022_Stiefel} show that Quasimatrix QR factorization is a retraction on our infinite dimensional Stiefel manifold $S_K$, so it follows that we may simply use the Quasimatrix version of QR factorization as our approximation to the exponential map in our updated algorithm.

\begin{algorithm*}
    \caption{Interpolation of continuous EGFs to unseen modeled parameters}
    \label{interpolation-scheme}
    \begin{algorithmic}
        \State \textsl{Input:} A set of singular functions and values, $\{[\mathcal{U}(s;\theta_j), \Sigma(\theta_j), \mathcal{V}(x;\theta_j),\theta_j]\}_{j=1}^{N_{\text{models}}}$ to be interpolated for a new parameter instance, $\theta_*$.
        \State \textsl{Step 1:} From among the $\theta_j$'s, identify a $\theta_0$, as being closest to $\theta_*$, and use the associated left singular functions, $\mathcal{U}(s;\theta_0)$, as the reference basis for the interpolation. Similarly, $\mathcal{V}(x;\theta_0)$ is the reference basis for interpolating the right singular functions.
        \State \textsl{Step 2:} Correct sign flips and shuffling of both left and right singular functions according to the left singular functions, $\mathcal{U}(s;\theta_0))$.
        \State \textsl{Step 3:} Perform the interpolation:
        \begin{enumerate}[noitemsep,nolistsep]
            \item Lift the left and right singular functions to the tangent spaces $\mathcal{T}_{\mathcal{U}_0} S_K$ and $\mathcal{T}_{\mathcal{V}_0} S_K$ respectively, by using the following map,
        
            \begin{align*}
                \Gamma^{\mathcal{U}_0}_j &= \mathcal{U}(s;\theta_j) - \mathcal{U}(s;\theta_0)\ \text{sym}\left(\mathcal{U}^{*}(s;\theta_0) \ \mathcal{U}(s;\theta_j)\right), \\
                \Gamma^{\mathcal{V}_0}_j &= \mathcal{V}(x;\theta_j) - \mathcal{V}(x;\theta_0)\ \text{sym}\left(\mathcal{V}^{*}(x;\theta_0) \ \mathcal{V}(x;\theta_j)\right), \quad \text{sym}\left(Y\right) \coloneqq (Y + Y^*)/2.
            \end{align*}
            
            \item Using Lagrange polynomials, compute the interpolated tangent quasimatrices, $\Gamma^{\mathcal{U}}_*$ and $\Gamma^{\mathcal{V}}_*$, using $ \{[\Gamma^{\mathcal{U}}_j, \Gamma^{\mathcal{U}}_j]\}_{j = 1}^{N_{models}}$. Interpolate the coefficients, $ \{\tilde{\Sigma}(\theta_j)\}_{j = 1}^{N_{models}}$, with the same scheme, to obtain $\Sigma(\theta_*)$.
            
            \item Compute the interpolated left and right singular functions, $\mathcal{U}(s;\theta_*), \mathcal{V}(x;\theta_*)$, by mapping $\bm{\Gamma}_*$ back to $\mathcal{S}_{K}$ using the exponential map,
        
            \begin{align*}
                \mathcal{U}(s;\theta_*) = \text{qf}(\mathcal{U}(s;\theta_0) + \Gamma^{\mathcal{U}}_*), \\
                \mathcal{V}(x;\theta_*) = \text{qf}(\mathcal{V}(x;\theta_0) + \Gamma^{\mathcal{V}}_*),
            \end{align*}
            where $\text{qf}(A)$ denotes the $Q$ factor of the QR decomposition of $A \in \mathbb{R}^{\infty \times r}$.
        \end{enumerate}
        \State \textsl{Step 4:} Order the singular functions and values to match the left singular functions at the initial parameter instance, $\theta_0$.
        \State \textsl{Output:} Interpolated left singular functions, $\mathcal{U}(s;\theta_*)$, right singular functions, $\mathcal{V}(x;\theta_*)$ and singular values $\Sigma(\theta_*)$ at parameter $\theta_*$.
    \end{algorithmic}
\end{algorithm*}

\begin{figure}[htbp]
\centering
\begin{overpic}[width=0.75\textwidth]{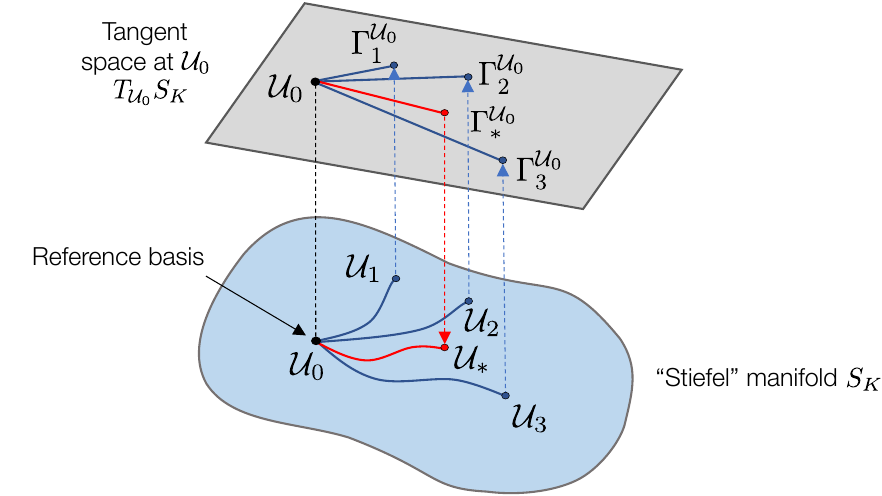}
\end{overpic}
\caption{The points on the ``Stiefel" manifold (represented by the orthonormal quasimatrices $\mathcal{U}_j$) are projected onto the flat tangent space $\smash{\mathcal{T}_{\mathcal{U}_0} S_K}$ at the reference basis $\smash{\mathcal{U}_0 = \mathcal{U}(s;\theta_0)}$, where they are interpolated and then returned to the compact Stiefel manifold as $\mathcal{U}_*$.}
\label{fig:Interp}
\end{figure}

\cref{fig:Interp} depicts a schematic of the described interpolation method from \cref{interpolation-scheme}.

\subsubsection{Correcting sign and order for singular functions}

\begin{figure}[htbp]
\centering
\begin{overpic}[width=\textwidth]{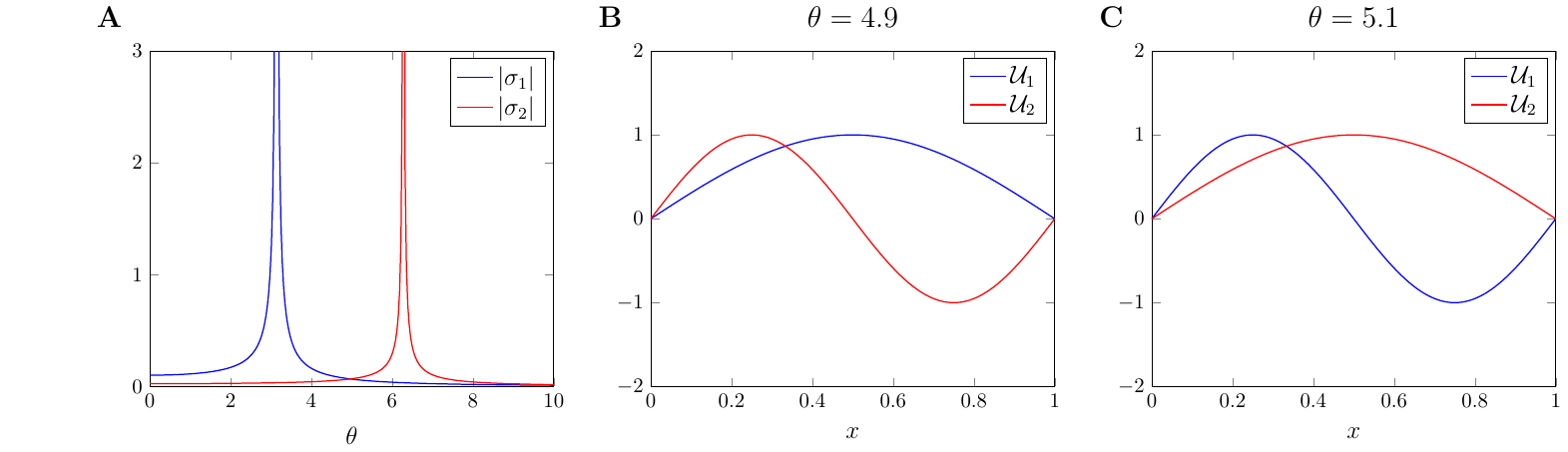}
\end{overpic}
\caption{(A) Magnitude of the first two singular values of the Green's function associated with the one-dimensional Helmholtz equation. The first two corresponding left singular functions are swapped when the frequency, $\theta$, increases beyond the critical value of $\theta_{\text{crit}} = \sqrt{5/2}\pi\approx 5$, as displayed in (B) and (C).}
\label{fig:mode-swap}
\end{figure}

The approximation for the Green's function, $\mathcal{G}_{\theta}(x,s) \equiv \mathcal{U} (y; \theta) \Sigma(\theta) \mathcal{V}^*(x;\theta)$, constructed in \cref{sec:chebgreen-construction}, is stored as a singular value expansion within a library of other such expansions, one each for the collection of model parameters, $\theta$, considered. As singular value expansion orders the singular functions according to the magnitude of the singular values by construction, the singular functions may swap their order as the parameter $\theta$ varies. If we consider the one-dimensional Helmholtz equation with frequency $\theta\geq 0$ and homogeneous boundary conditions:
\[\frac{d^2u}{dx^2}+\theta^2 u=f,\quad x\in[0,1], \quad \text{or} \quad \mathcal{L}_\theta = \frac{d^2}{dx^2}+\theta^2.\]
We observe that the first two singular functions swap at some critical value of $\theta$. The $k^{\text{th}}$ singular value for the Green's functions of the operator $\mathcal{L}_\theta$ are given by $\sigma_k = 1/(\theta^2 - (k\pi)^2)$. If we compare the first two modes,  they satisfy $|\sigma_1|>|\sigma_2|$, when $\theta<\sqrt{5/2}\pi$, and $|\sigma_1|\leq|\sigma_2|$ otherwise. This implies that the first two left singular functions are swapped when $\theta>\sqrt{5/2}\pi$, as illustrated by \cref{fig:mode-swap}, and that the manifold interpolation technique will perform poorly in such cases. The same phenomena occurs for the higher singular functions at larger values of $\theta$. We propose to reorder the singular functions and the associated singular values of the discovered basis at the given interpolation parameter to match the ones at the origin point $\theta_0$, where we lift to the tangent space of the infinite-dimensional analogue for Stiefel manifold. For a given parameter, $\theta_j$, and mode number, $1\leq k\leq K$, we select the $k^{\text{th}}$ left singular function at $\theta_j$ to be the one with minimal angle with the $k^{\text{th}}$ left singular function at parameter $\theta_0$, \emph{i.e.},
\begin{equation*}
    \begin{aligned}
        &\mathcal{U}_k (s; \theta_j) = \mathcal{U}_{\ell'} (s; \theta_j),    \quad \text{where}\quad \ell' = \text{match}(k), \\
        \text{match}(k) = &\argmax_{1\leq \ell\leq K, \ell \ne \text{match}(k') \forall 1 \le k' < k} |\langle \mathcal{U}_{\ell} (s; \theta_j), \mathcal{U}_{k} (s; \theta_0) \rangle|,\quad 1\leq k\leq K.  
    \end{aligned}  
\end{equation*}
where $\langle\cdot,\cdot\rangle$ denotes the inner product in the continuous $L^2$ sense. Note that the reordering is done by matching the left singular functions, starting from the left singular function with the highest singular value, without replacement. When the order of the left singular functions has been changed by this procedure, we also re-order the corresponding singular values and right singular functions to preserve the value of the Green's function.

 Another potential issue arises because singular functions are unique only up to a sign flip. Thus, learning approximations to Green's functions at two different parameters close to each other, $\theta_1$ and $\theta_2$ say, might lead to singular functions associated with the $k^{\text{th}}$ singular value being very ``dissimilar" between the Green's functions learned at the respective parameter value, $\theta$. In order to best align the singular functions before interpolation, we account for these sign flip by comparing the the left and right singular functions of the learned approximations with an inner product. Consider the left singular functions at the origin of the manifold, $\mathcal{U}_k (s;\theta_0))$ (implying the $k^{\text{th}}$ left singular function), around which we form the tangent space. We compute an inner product between the left singular functions of the other interpolants, $\mathcal{U}_k(s;\theta_j))$, and the corresponding left singular functions at the origin, $\mathcal{U}_k(s;\theta_0)$, in order to correct the signs for the left and right singular functions of the interpolant, as follows:
\[\mathcal{U}_k(s;\theta_j), \mathcal{V}_k(x;\theta_j) \equiv 
\begin{cases}
-\mathcal{U}_k(s;\theta_j), - \mathcal{V}_k(x;\theta_j)& \text{if}\ \langle \mathcal{U}_k(s;\theta_j), \mathcal{U}_k(s;\theta_0) \rangle < 0,\\
\mathcal{U}_k(s;\theta_j), \mathcal{V}_k(x;\theta_j) &\text{otherwise}.
\end{cases}\]

Note that the reshuffling of the singular functions is performed before correcting for the sign flips as we are making the assumption that the singular functions associated with the $k^{\text{th}}$ singular value across models are somewhat correlated.

\subsection{A note on computational complexity}
The following subsections provide a note on the computational complexity of all the key operations used in the \verb|chebgreen| package:

\cprotect\subsubsection{\verb|chebfun2|}
Constructing a \verb|chebfun2| for a bivariate function $g(x,y)$ (\cref{subsec:chebfun2-aprox}) requires sampling the function on an $N \times N$ Chebyshev tensor grid. $N$ depends on the accuracy prescribed in $x$ and $y$ directions. This results in a matrix of sampled values. The process then involves computing an approximate singular value decomposition (SVD) of this matrix. To obtain a near-optimal rank $K$ approximation, the computational complexity is $\mathcal{O}(K^2 N + K^3)$ \cite{Townsend_Chebfun2}.

\subsubsection{QR decomposition}
The implementation for a QR decomposition of a column Quasimatrix, $\mathcal{A}$, of size $(\infty \times K)$ (\cref{sec-singular-value-expansion}), where $K$ is the rank of the column Quasimatrix, is analogous to the Householder QR decomposition for a conventional matrix when working in a space with an $L^2$ inner product. Since the columns of a column Quasimatrix are represented by a \verb|chebfun|, the additional cost is the conversion between the Chebyshev coefficients, which represent the \verb|chebfun| and the corresponding values of the function at Chebyshev points. This conversion utilizes the Fast Fourier Transform (FFT) for coefficients to values, and the Inverse Fast Fourier Transform (IFFT) for values to coefficients. Assuming each chebfun in $\mathcal{A}$ is represented by $M$ coefficients, the cost for these transforms is $\mathcal{O} (K M log(M))$. This cost is dominated by the cost of computing the matrix Householder QR decomposition, which has a complexity $\mathcal{O} (d^2 D)$, where $d = \min (M,K)$ and $D = \max(M,K)$. Given that generally $K < M$, the total computational complexity for the QR decomposition is $\mathcal{O} (K^2 M)$.

\subsection{Singular Value Expansion}
A similar argument is valid for computing the Singular Value Expansion (SVE) of a \verb|chebfun2|, which is internally stored as a CDR decomposition. Let $C$ be a $(\infty \times K)$ Quasimatrix, $R$ be a $(L \times \infty)$ Quasimatrix, and $D$ be a $(K \times L)$ matrix. Assume $M$ is the number of coefficients used to represent the \verb|chebfun|-s in both $C$ and $R$. Without loss of generality, assuming $K > L$, the dominant cost in computing the SVE arises from the QR decompositions performed in \cref{sve-chebfun2}: Step 1. Hence, the computational complexity for computing the SVE is $\mathcal{O}(K^2 M)$.

% - SVD: 
% Assuming a CDR decomposition C - (inf, K), D - (K,L) and R - (L, inf)
% Assuming $K > L$, without loss of generality:

% QR: O($K^2 M$) + O($L^2 M$) = O($M(K^2 + L^2)$) = O($M K^2$)
% Matrix multiply: O($K^2 L + K L^2$) = O($K L (K + L)$) = O ($K^3$)
% Matrix SVD: O($K L^2$) if $K > L$ or O($L K^2$) if $L > K$ ~ O($K^3)$
% Compute Quasimatrix matrix product: O($M K^2$) + O($L^2 M$) ~ O($K^2 M$)

\subsubsection{Interpolation}
For interpolating between models of Green's functions, the primary computational cost lies in lifting the left and right singular functions for all interpolant models to the tangent space at the reference basis (Section \ref{interpolation-scheme}: Step 1). Assuming that each interpolant model has a rank $K$ singular value expansion and that the \verb|chebfun|-s representing the left and right singular functions are represented by $M$ coefficients, the computational complexity for interpolating between $N_{\text{models}}$ interpolant models is  $\mathcal{O} \left( N_{\text{models}} K^2 M\right)$.

\section{Theoretical error analysis} \label{sec:error-analysis}
In this section we consider each of the main sources of error for our interpolation approach: that from the orthogonal projection onto the tangent space of the Hilbert Stiefel manifold \(S_K\); that from the Lagrange polynomial interpolation of quasimatrices in the tangent space; and, that from the \(QR\)-based retraction back to the manifold. Although the error analysis for the Lagrange polynomial interpolation is a classical result, little attention has been given to error analysis for the other two steps in general, but especially in the Hilbert manifold setting. Furthermore, no attention has been given to quantifying the size of the region about a base point by which these methods are valid. The proofs for the results presented in this section are given in Appendix B.

\subsection{Error due to the projection step} The projection mapping computed in Theorem 1 is the first instance in Algorithm 1 of introducing approximation error after the Green's functions are represented as chebfuns. Exploiting properties of the Hilbert Stiefel manifold \(S_K\), we demonstrate that the error in this step decays at least cubically as the distance between points approaches zero in a suitable region of the domain. Before we do this, however, we establish a concrete region within which the Riemannian exponential map is guaranteed to be a local diffeomorphism: this quantity is known as the \emph{radius of injectivity}.

\begin{lemma}
    The radius of injectivity of the Riemannian exponential map on the Hilbert Stiefel manifold \(S_K\) of \( H = (L^2(\Omega))^K\) is \(\pi\).
\end{lemma}

This result tells us that, if \(\Phi\) is our base point and \(\Psi\) another point of interest, then the the map \(\exp_{\Phi}\colon T_{\Phi}S_K\rightarrow S_K\) is a local diffeomorphism, say on some neighborhood \(U\) of \(\Phi\) in \(T_{\Phi}S_K\) whose image is a geoedsic ball of at most radius \(\pi\) in the geodesic distance on \(S_K\). While the geodesic distance may seem to be an abstract way to quantify this region, given that our method does not use this distance directly, however it does furnish an analogous result for the \(H\)-norm on a tangent vector: that is if \(\Gamma = \exp_{\Phi}(\Psi)\), then the quantity \(\| \Gamma\|_H\) is equal to the geodesic distance between \(\Phi\) and \(\Psi\) \cite{stoye2024injectivityradiusstiefelmanifold}.

Now that we have quantified a region on which the exponential map is a diffeomorphism, we may begin quantifying error estimates, starting with that which is due to the orthogonal projection on \(S_K\).

\begin{proposition}[Cubic approximation error of the projection mapping onto the tangent space]
Let \(S_K\) be the Hilbert-Stiefel manifold endowed with the Riemannian metric induced from
\[
H = \bigl(L^2(\Omega)\bigr)^K,
\]
and let \(\phi \in S_K\). Suppose the neighborhood \(U\) of the origin in \(T_\phi S_K\) contains only tangent vectors \(v\) such that \(\|v\|_H\ < \pi \). Let \(P_\phi: H \to T_\phi S_K\) be the orthogonal projection onto \(T_\phi S_K\) . Then there exists a constant \(C>0\) such that, for every \(y \in W\) with \(y=\exp_\phi(v)\), where \(W\) is the image of the exponential map on \(U\), we have
\[
\|v-P_\phi(y-\phi)\|_H\le C\,\|y-\phi\|_H^3.
\]
\end{proposition}

This is a somewhat surprising result. One might assume that, doing a Taylor expansion in order to compute the error between a projected point and an actual point that error would be quadratic in the distance between points as this distance approaches zero. But it turns out that, even in the Hilbert manifold case \cite{biliotti2006exponentialmapweakriemannian}, in a region where the Riemannain exponential map is a local diffeomorphism, the second order term of the Taylor expansion of the exponential map is identically zero \cite{Boumal_2023}. This is the key to obtaining the above result.

\subsection{Error due to the \(QR\)-based retraction}
After interpolating tangent vectors in the tangent space, we map back down to the Hilbert Stiefel manifold using a mapping that sends a quasimatrix to the \(Q\) in its \(QR\) factorization. This step represents the bulk of the error in the entire interpolation process, dominating the error in both the projection and Lagrange polynomial interpolation steps. The following result is well-established in the finite dimensional case, and is a consequence Taylor's theorem for Banach spaces.

\begin{proposition}[Quadratic error for a retraction]
Let \(M\subset H\) be an embedded Hilbert manifold in a Hilbert space \(H\), so that each tangent space \(T_{\phi}M\) is a linear subspace of \(H\). Suppose that 
\[
R: TM \to M,\quad (\phi,\eta) \mapsto R(\phi,\eta)
\]
is a \(C^2\)  retraction, that is that:
\begin{enumerate}
  \item[(i)] \(R(\phi,0)=\phi\).
  \item[(ii)] \(DR(\phi,0)[\eta]=\eta\) for all \(\eta\in T_{\phi}M\).
  \item[(iii)] \(R(\phi,\cdot)\) is \(C^2\). Hence, by continuity there exists a constant \(L>0\) (depending on \(\epsilon\)) such that
  \[
  \|D^2R(x,0)[\eta,\eta]\|_H \le L\,\|\eta\|_H^2\quad\text{for all } \eta\in B_\epsilon(0).
  \]
\end{enumerate}
Then for every \(x\in M\) and every \(\eta\in T_xM\) with \(\|\eta\|_H<\epsilon\), there exists \(\delta>0\) such that for all \(t\) with \(|t|<\delta\) we have
\[
\|R(x,t\eta) - (x+t\eta)\|_H \le L\,t^2\,\|\eta\|_H^2.
\]
\end{proposition}
We now must clarify that the \(QR\)-based retraction is \(C^2\). In fact, it is smooth.

\begin{lemma}[Smoothness of the Gram--Schmidt map on an open subset of $H$]
Let
\[
H = \left( L^2(\Omega) \right)^K
\]
be a Hilbert space and let $M$
be the Hilbert Stiefel manifold of $H$. Define the map
\[
qf: U\subset H \to M,
\]
by applying the Gram--Schmidt process, that is, define
\[
q_1(\phi_1) = \frac{\phi_1}{\|\phi_1\|_H},
\]
and for \(j=2,\dots,K\), recursively set
\[
p_j(\phi_1, \dots, \phi_j) = \phi_j - \sum_{i=1}^{j-1} \langle \phi_j, q_i(\phi_1,\dots,\phi_i) \rangle_H\, q_i(\phi_1,\dots,\phi_i),
\]
\[
q_j(\phi_1,\dots,\phi_j) = \frac{p_j(\phi_1,\dots,\phi_j)}{\|p_j(\phi_1,\dots,\phi_j)\|_H}.
\]
Then, the mapping
\[
qf(\phi) = \bigl( q_1(\phi_1), \; q_2(\phi_1,\phi_2), \; \dots, \; q_K(\phi_1,\dots,\phi_K) \bigr)
\]
is smooth on the open set
\[
U = \{ \phi \in H : \|\phi_i\|_H > 0 \text{ for } i=1,\dots, K, \text{ and } \{\phi_1,\dots,\phi_K\} \text{ is linearly independent} \}.
\]
\end{lemma}

It follows that the error associated with the \(QR\)-based retraction employed in our algorithm decays quadratically as the norm distance between the quasimatrices approaches zero.

\subsection{Combined error}
Now we consider the total error associated with the interpolation scheme. In the projection step, we obtain cubic order error. After this step, we use Lagrange polynomial interpolation, which has error of order \(\mathcal{O}(h^{n+1})\), where \(n\) is the number of interpolants and \(h\) the step size. Finally, we get quadratic error in the retraction mapping back to the manifold, which dominates the error of the other two steps in the interpolation scheme.

\section{Numerical results} \label{sec:numerical-examples}

In this section, we exemplify the use of our methods on a number of synthetic problems. Unless specified otherwise, the hyperparameters involved in generating the dataset, learning and storing the Green's function, and interpolating the same are as follows:

\begin{itemize}
    \item \textit{Generating the dataset}: The length scale of the squared exponential covariance kernel in the Gaussian Process from which the forcing terms are sampled is set to $\sigma = 10^{-2}$. The forcing terms and system responses are measured on a uniform 1D grid at $N_{\text{x}}= N_{\text{s}} = 500$ equally spaced points. It is crucial that the density of sampling points is chosen so that functions are sufficiently resolved on the grid, \emph{i.e.}, and the discrete vector of values at the sampling locations are free from spatial aliasing. In our case, since the forcing terms and responses are sampled from a continuous representation in a Chebyshev basis (using \verb|chebfun|), we use the number of Chebyshev points used to represent these functions as a proxy for the number of samples needed on a uniform grid. Note that equispaced points suffer from Runge Phenomenon \cite{Trefethen_Runge} so a uniform grid is not optimal for representing these functions but data are generally collected at uniform grids in experiments; thus we chose to test our methods in a such a setting. Finally, the number of input-output pairs is set to $N_{\text{samples}} = 100$, out of which only $~95\%$ are used during training. The choice of $100$ samples is motivated by a study on the fidelity in learning the Green's function for the Laplacian operator in \cref{subsubsec:error-samples}.
    
    \item \textit{Greenlearning}: Each of the neural network used to learn the Green's function and the homogeneous solution are $4$-layer Rational Neural Networks having $50$ neurons within each layer. They are trained for $2000$ epochs using the Adam optimizer with a learning rate of $10^{-2}$, and the learning rate decaying exponentially to $10^{-3}$ over the $2000$ epochs.
    
    \item \verb|chebpy2|: When learning polynomial approximations for one and two-dimensional functions using \verb|chebpy2|, we specify the accuracy of the functions that need to be resolved in the domain as $\varepsilon^{\text{cheb}} \approx 2.22 \times 10^{-16}$. In case of two dimensions, we specify this value for each of the independent variable within the domain, $\varepsilon^{\text{cheb}}_x$ and $\varepsilon^{\text{cheb}}_s$ for $G(x,s)$. These are both set to $\varepsilon^{\text{cheb}}_x = \varepsilon^{\text{cheb}}_s \approx 2.22 \times 10^{-16}$. We choose the value of $\approx 2.22 \times 10^{-16}$, which is the \textit{eps} - the difference between $1.0$ and the next smallest representable floating point number larger than $1.0$ for single floating point precision, in a direction for all experiments as our neural networks are trained with \verb|float64| (double floating point) precision. This choice affects the rank $K$ of the learned Green's function approximation. The rank $K$, defined as the number of rank $1$ functions in the bivariate representation in \cref{eq: chebfun2-approx}, is determined adaptively and depends on the specified error tolerances; $K$ increases as $\varepsilon^\text{cheb}_x$ and $\varepsilon^\text{cheb}_s$ are reduced for finer resolution. In the cases where there is noise in the collected data, as is the case when data are collected from experiments, this number can be set close to the noise floor so that the model does not overfit to noise. However, in our numerical experiments with noise, we do not change this number, so as to demonstrate the robustness of our method to noise.
\end{itemize}

For doing experiments with noisy dataset, we artificially pollute the system responses, $u_i$, with additive, white Gaussian noise as:
    \begin{equation*}
    u^{\text{noisy}}_i(x_j) = u_i(x_j) + \zeta c_{i,j} \overline{|u_i|}, \quad 1\leq i\leq N_{\text{sensors}},
    \end{equation*}
where $\overline{|u_i|}$ is the average of the absolute value of the $i$th system's response and the $c_{i,j}\sim \mathcal{N}(0,1)$ are independent and identically distributed. The noise level is controlled by the parameter $\zeta$ which we have set to $\zeta = 0.5$, which corresponds to $50\%$ noise in the output.

In the cases, where we know the analytical Green's function for our numerical experiments, we can assess the learned Green's function in an $L^2$-sense, we define the ``relative error" as:
    \begin{equation}
        \epsilon = \dfrac{\|\mathcal{G}(x,s) - G_{\text{exact}}(x,s)\|_{L^2(\Omega \times \Omega)}}{\|G_{\text{exact}}(x,s)\|_{L^2(\Omega \times \Omega)}}
        \label{eq:analytical-relative-error-metric}
    \end{equation}
where, $\mathcal{G}(x,s)$ is the \verb|chebgreen| model of the Green's function and $G_{\text{exact}}(x,s)$ is the closed-form Green's function of the underlying operator. However, this is not possible in all cases since the closed-form for the Green's function is not known in all cases. In these cases, we propose that one can compute a ``test error" by reconstructing the system responses for unseen data (the $5\%$ of the dataset we leave out for validation during training). For this paper, we generate a testing dataset with $N_{\text{test}} = 100$ pairs of forcing functions and system responses, $\{f_i, u_i\}_{i = 1}^{N_{\text{test}}}$ and compute the test error as:

\begin{equation}
    \epsilon_{\text{test}} = \dfrac{1}{N_{\text{test}}} \sum_{i=1}^{N_{\text{test}}} \dfrac{\|\tilde{u}_i - u_i\|_{L^2(\Omega)}}{\|u_i\|_{L^2(\Omega)}} \times 100 \%, \quad \tilde{u}_i = \int_\Omega \mathcal{G}(x,s)f_i(s) ds + \mathcal{H}(x).
    \label{eq:relative_error_metric}
\end{equation}

In the next part of this section, we limit discussion to cases where the homogeneous solution, $\mathcal{H}(x)$, lives within a sub-space to the given problem’s solution space (e.g. homogeneous Dirichlet and periodic cases).

\subsection{Poisson problem}
    Before we demonstrate our method's efficacy for interpolation, we will approximate the Green's function, and its singular value expansion, for a one-dimensional Laplacian operator. Assume we have a Poisson problem with homogeneous boundary conditions defined as follows:

    \begin{equation}
        -\frac{d^2 }{d x^2} u(x) = f(x), \quad x \in [0,1], \quad u(0) = u(1) = 0.
        \label{eq:laplace}
    \end{equation}

    For this is canonical problem, the closed-form solution for the associated Green's function is known and is given by:

    \begin{equation*}
        G_{\text{exact}} = \begin{cases}
                            x (1 - s), \quad \text{if } x \le s, \\
                            s (1 - x), \quad \text{if } x > x,
                            \end{cases}
    \end{equation*}
    where $x,s \in [0,1]$.

\begin{figure}[htbp]
        \centering
        \begin{overpic}[width=\textwidth]{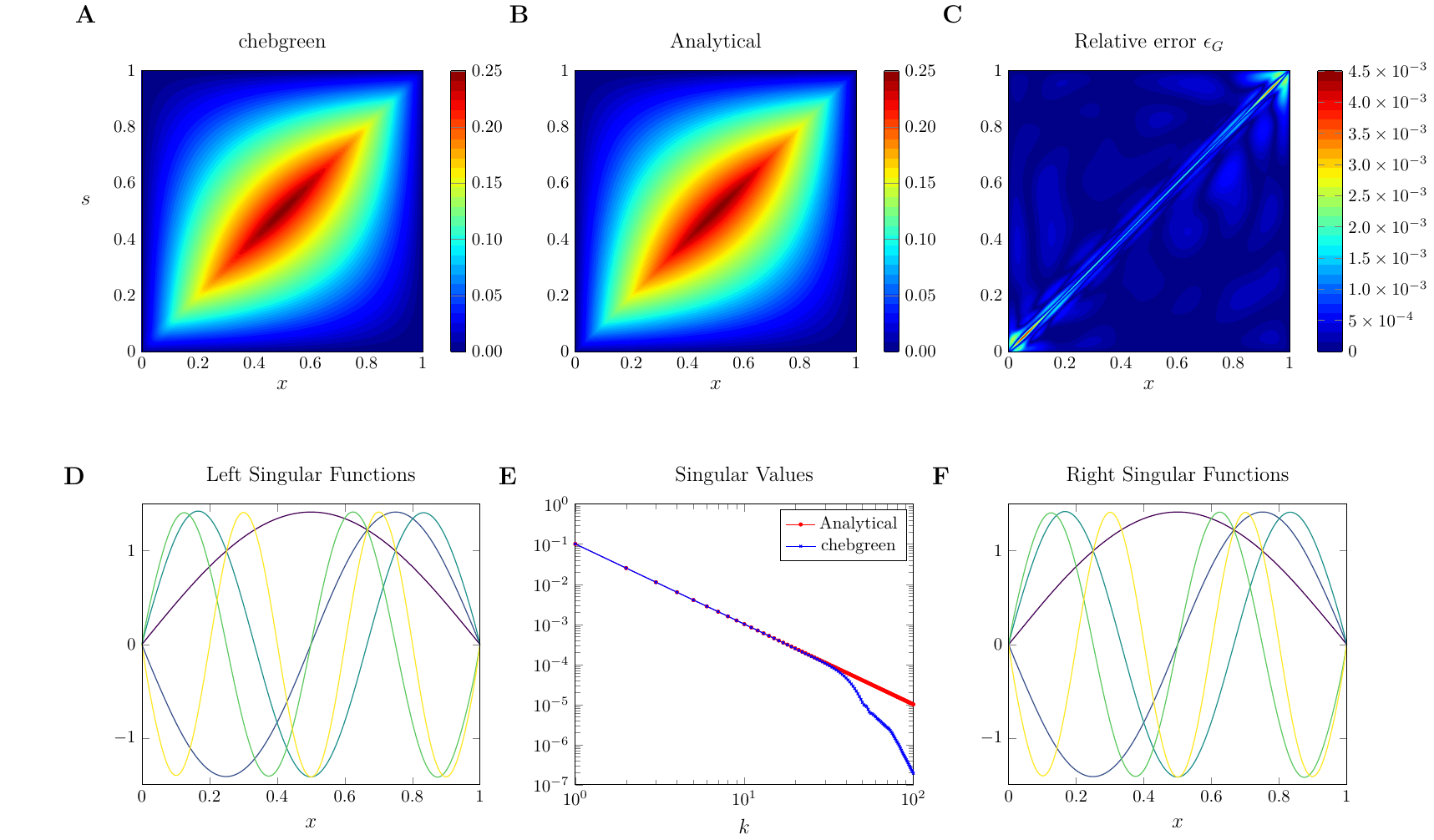}
        \end{overpic}
        \cprotect\caption{(A) Approximation of the Green's functions learned by \verb|chebgreen|, (B) Analytical solution of the Green's function associated with the Laplacian operator, (C) Relative error contour between the learned model and analytical solution, (D) First 5 Left Singular Functions of the learned Green's function, (E) Comparison of the learned singular values against the analytical singular values for the Green's function, and (F) First 5 Right Singular Functions of the learned Green's function.}
        \label{fig:laplace}
    \end{figure}

    Using \verb|chebgreen|, we learn the Green's function for the underlying Laplacian operator using noise-free data, as shown in \cref{fig:laplace}. On visual inspection, the Green's function learned by \verb|chebgreen| (\cref{fig:laplace}A) matches closely with the analytical Green's function for \cref{eq:laplace} (\cref{fig:laplace}B). In \cref{fig:laplace}C, we plot the relative error between the learned and analytical Green's function,
    \begin{equation*}
        \epsilon_G = \dfrac{|\mathcal{G}(x,s) - G_{\text{analytical}}(x,s)|}{\|G_{\text{analytical}}(x,s)\|_{L^2([0,1] \times [0,1])}}
    \end{equation*}
    as an error contour. The relative error $\epsilon = 0.36 \%$ indicates satisfactory the fidelity of our learned Green's function to the analytical solution.
    Finally, we observe that the first five learned left (\cref{fig:laplace}D) and right (\cref{fig:laplace}F) singular functions are indistinguishable from the left and right singular functions of the analytical solution of the Green's function. In \cref{fig:laplace}E, we plot the first 100 learned singular values for the Green's function against the analytical values of the singular values, $1/(\pi^2 k^2)$. There is an exponential decay for the smallest singular values in the case of the learned Green's function. We use Rational NN to approximate the Green's function which is a smooth approximation to the exact Green's function; thus, this fast decay is expected. The key point is that the largest learned singular values are in close agreement with the analytical solution. Therefore, our method constructs a high-fidelity low-rank approximation of the Green's function associated with the Laplacian operator.

    \subsubsection{Study on effects of number of samples, $N_{\text{samples}}$} \label{subsubsec:error-samples}
    To determine an appropriate number of samples, $N_{\text{samples}}$,  for achieving sufficient fidelity to the solution operator, we conducted a study examining the relationship between the number of samples used and the resulting error. Our analysis revealed a significant reduction in error as the sample count increased, initially. However, this improvement exhibited diminishing returns, plateauing at approximately $100$ samples. As the theoretical results \cite{Boulle2021_Theory} suggest, for a covariance kernel of a given quality, only a relatively small degree of oversampling is necessary to learn a rank $K$ approximation to underlying Green's function. Consequently, the plateau in error reduction observed around $100$ samples likely indicates that we have extracted the majority of the learnable information about the underlying solution operator, given the characteristics of our GP kernel and the quality of our training data. Therefore, we proceeded with 100 samples in our subsequent numerical experiments, balancing computational cost with achieving a level of fidelity consistent with the theoretical limits.

    \begin{figure}[htbp]
        \centering
        \begin{overpic}[width=\textwidth]{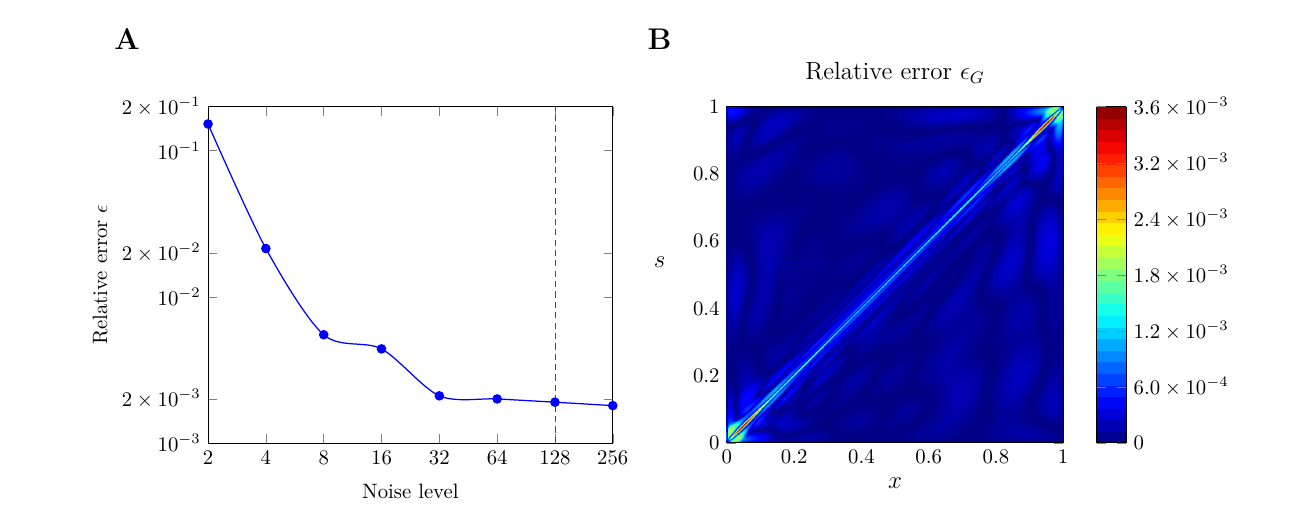}
        \end{overpic}
        \caption{(A) Relative error for the Laplacian operator when trained with different number of samples, $N_{\text{train}}$, (B) Contours for relative error between the learned model with a noisy dataset at $N_{\text{train}} = 128$ and analytical solution for the Laplacian operator.}
        \label{fig:laplace_sample_study}
    \end{figure}

    \subsubsection{Study on effects of noise in the samples}
\begin{figure}[!htbp]
        \centering
        \begin{overpic}[width=\textwidth]{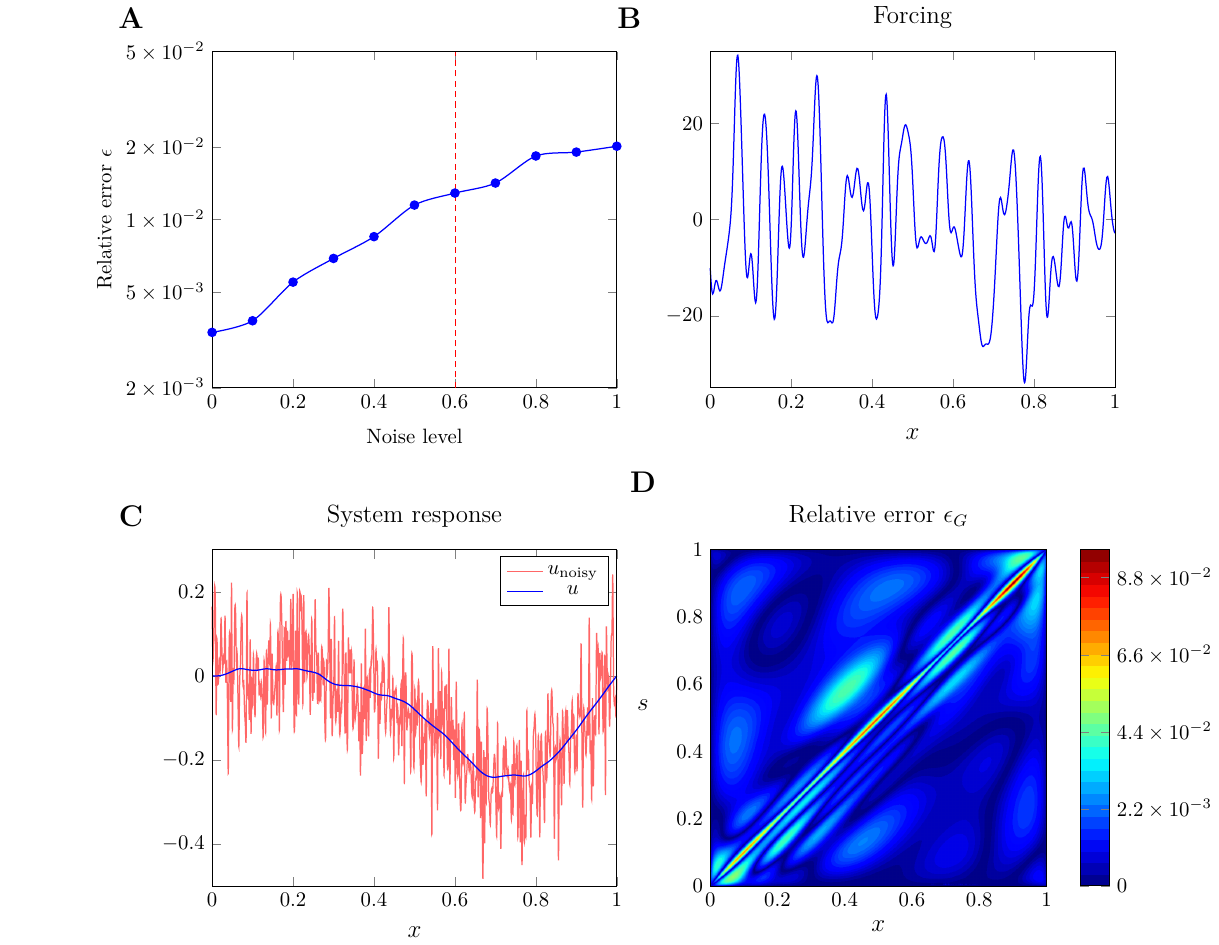}
        \end{overpic}
        \cprotect\caption{(A) Relative error at different noise level, $\zeta$, for the Laplacian operator, (B) Instance of a forcing function, (C) Instance of a system response with and without white Gaussian noise at level $\zeta = 0.6$ ($60\%$ noise), and (D) Contours for relative error between the learned model with a noisy dataset at $\zeta = 0.6$ and analytical solution for the Laplacian operator.}
        \label{fig:laplace_noise_study}
    \end{figure}

    \cref{fig:laplace_noise_study}A shows how the relative error for the learned Green's function, $\epsilon$, changes as a function of the noise level, $\zeta$, for the system responses in the training dataset. Note that we only consider the effect of output noise in this case. An instance of a possible sample from the dataset with $\zeta = 0.6$ ($60 \%$ noise) is shown in \cref{fig:laplace_noise_study}: B is the forcing function used to perturb the system, and C is the corresponding clean system response (blue) along with the artificially polluted system response (red). Finally, the error contour of the Green's function learned with a dataset with noise level $\zeta = 0.6$ against the analytical Green's function is shown in \cref{fig:laplace_noise_study}D. The relative error is $\epsilon = 1.29 \%$. Thus, even when the system responses barely resemble the underlying function (because of noise), our method seems to works well; learning a reasonably accurate approximation to the underlying Green's function. 
    
\subsection{Advection-Diffusion}
    As the next problem, we learn and interpolate the Green's function associated with the operator in a parameterized Advection-Diffusion case, described by the following boundary value problem:
    \begin{align*}
        \frac{d^2 }{d x^2} u(x) + \theta \frac{d }{d x} &u(x)= f(x), \quad x \in [-1,1],\\
        &u(-1) = u(1) = 0.
    \end{align*}
    Thus, the operator for which we would like to find a Green's functions (in the domain $x \in [-1,1]$) is:

    \begin{equation*}
        \mathcal{L}_\theta = \frac{d^2 }{d x^2} + \theta \frac{d }{d x}
    \end{equation*}

    \begin{figure}[htbp]
        \centering
        \begin{overpic}[width=\textwidth]{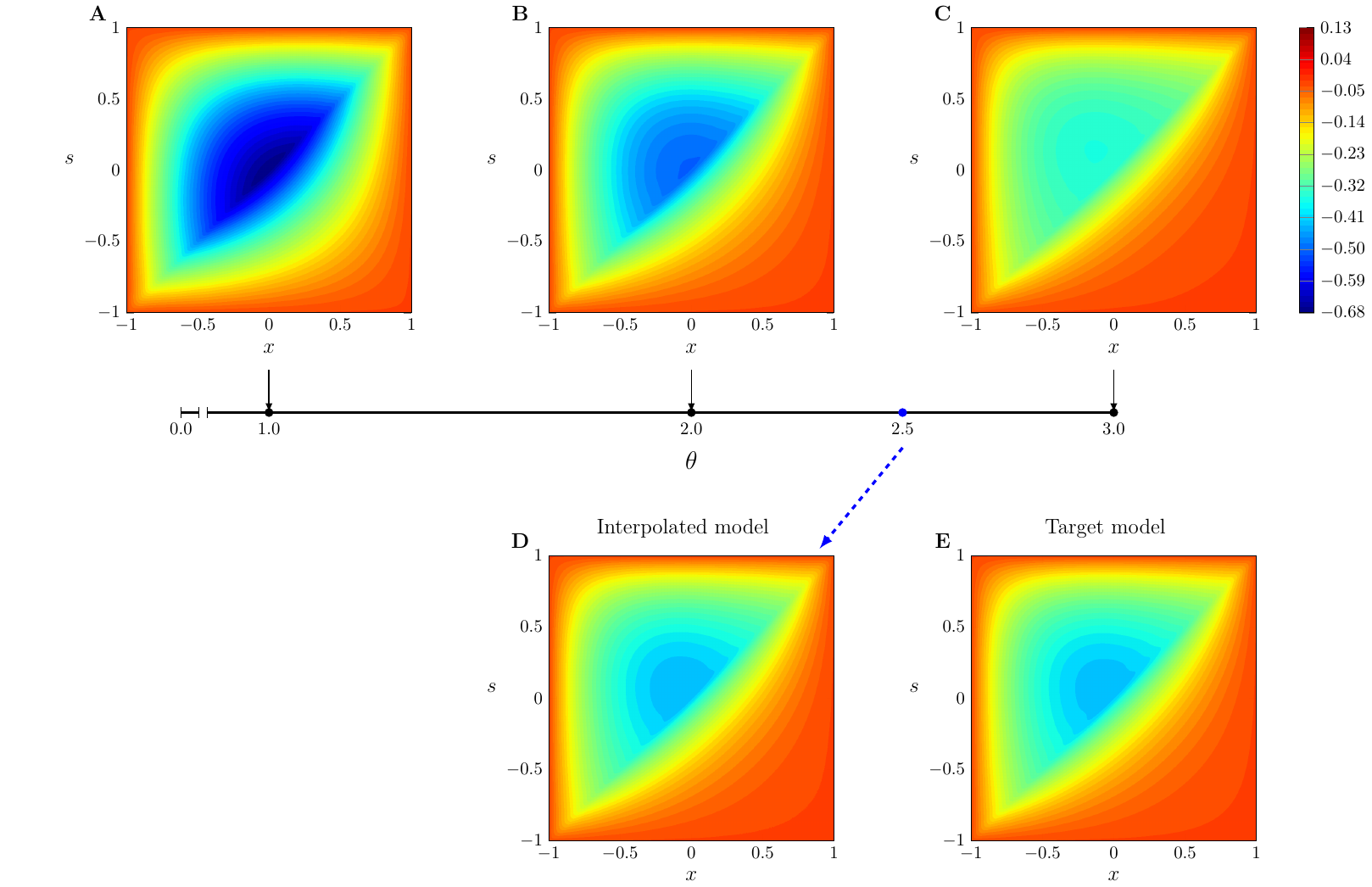}
        \end{overpic}
        \cprotect\caption{(A-C) Approximation of the Green's functions associated with the Advection-Diffusion problem at $\theta=1,2,3$ used by the interpolation scheme. (D) Interpolated Green's function at $\theta_*=2.5$ ($\epsilon_{\text{test}} = 0.54 \%$). (E) The target Green's function approximated by \verb|chebgreen|.}
        \label{advection_diffusion}
    \end{figure}

    We learn approximations to the Green's functions for the associated non self-adjoint linear operator at the parameter values $\theta_1 = 1.0$, $\theta_2 = 2.0$, and $\theta_3 = 3.0$. The test error for the approximations is less than $0.29 \%$; computed using our small training set with $N_{\text{train}} = 237$ input-output pairs for each $\theta$.
    
    Using these three interpolants, we compute an interpolated model at $\theta_* = 2.5$. We also compute a target model (ground truth), or a model compute with data generated at $\theta_* = 2.5$ for comparison. As we can see in \cref{advection_diffusion}, the interpolated and the target Green's functions are in close agreement, on visual inspection. The test error for the interpolated Green's function at $\theta_*$ is equal to $\epsilon_{\text{test}} = 0.54 \%$. In this case, we can also compare the interpolated Green's function against the analytical solution by computing the relative error, which is $\epsilon = 0.26 \%$. This demonstrates that our model can approximate Green's functions for non self-adjoint linear operators extremely well with a small amount of data as well as the manifold interpolation technique provides a great framework to interpolate these learned models based on some parameter value.

    \begin{figure}[htbp]
        \centering
        \begin{overpic}[width=0.8\textwidth]{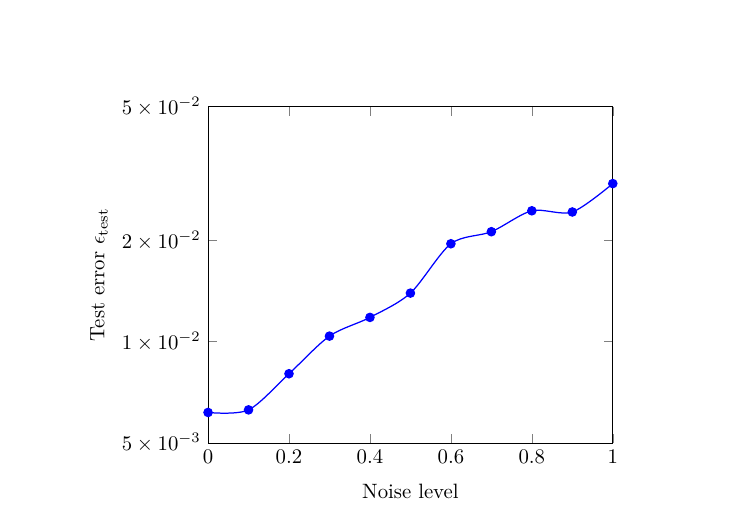}
        \end{overpic}
        \caption{Relative error at different noise level, $\zeta$, for the interpolated Advection-Diffusion operator at $\theta = 2.5$.}
        \label{fig:interp-error}
    \end{figure}

    We also show that our interpolated model is robust to noise in the training data used to create the individual interpolants (\cref{fig:interp-error}). This figure illustrates how the test error ($\epsilon_{\text{test}}$) of the interpolated Green's function changes as the output noise level ($\zeta$) increases in the system response data used to train the interpolants. Note that even when the level of noise in the training data is $\zeta = 1.0$ ($100 \%$), the test error is still less than $5\%$.
    
\subsection{Airy Problem}
    After establishing the fidelity of the Green's function learned and interpolated by our method to the analytical expression for the same in cases where these were available, we now demonstrate \verb|chebgreen| on problems where the analytical form of the Green's function is not known. To this end, we parameterize the Airy equation in the following way:

    \begin{equation*}
        \frac{d^2u}{dx^2} - \theta^{2} x u = f, \quad x \in [0,1],
    \end{equation*}
    In this case, we demonstrate both interpolation and \textit{extrapolation} capabilities of \verb|chebgreen|:
\subsubsection{Interpolation}
\begin{figure}[htbp]
        \centering
        \begin{overpic}[width=\textwidth]{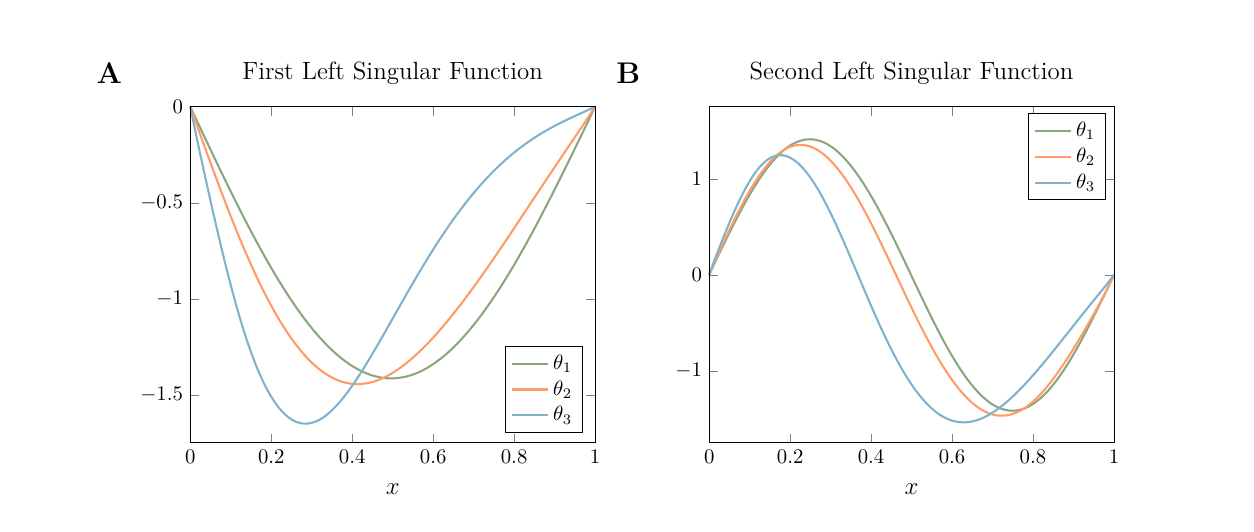}
        \end{overpic}
        \cprotect\caption{(A) First and (B) Second learned Left Singular Function for the airy problem at $\theta_1 = 1.0$, $\theta_2 = 5.0$, and $\theta_3 = 10.0$}
        \label{fig:airy_modes}
    \end{figure}

    For the interpolation case, we compute the three approximations to the Green's functions for the parameterized operator at $\theta_1 = 1.0$, $\theta_2 = 5.0$, and $\theta_3 = 10.0$ (we call these the ``interpolant" cases). The left and right singular functions for this problem vary significantly as we change the value of $\theta$ as seen in \cref{fig:airy_modes}, which makes it a useful test for our interpolation method.

\begin{figure}[htbp]
        \centering
        \begin{overpic}[width=\textwidth]{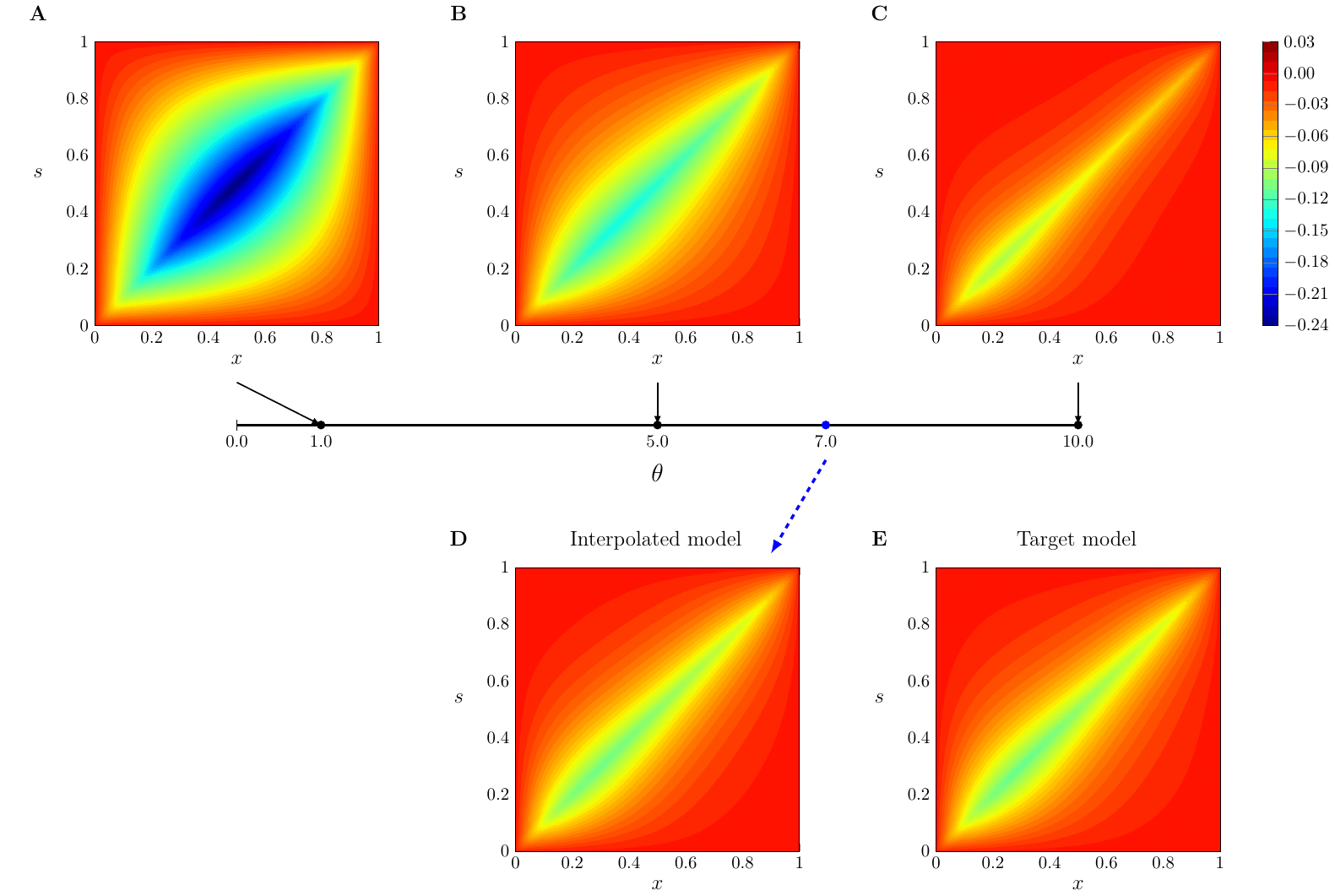}
        \end{overpic}
        \cprotect\caption{(A-C) Approximation of the Green's functions associated with the Airy problem at $\theta=1,5,10$ used by the interpolation scheme. (D) Interpolated Green's function at $\theta_*=7$ ($\epsilon_{\text{test}} = 2.76 \%$). (E) The target Green's function approximated by \verb|chebgreen|.}
        \label{airy_interpolation}
    \end{figure}

    Note that since the values of the parameter $\theta$ are much further apart compared to the Advection-Diffusion case, and $\theta$ appear as a squared quantities in the operator, we expect the differences in the learned Green's function to be much more drastic. This is evident from the learned Green's function shown in \cref{airy_interpolation}A-C. The interpolated Green's function at $\theta_*$ has a test error of $\epsilon_{\text{test}} = 2.76 \%$. This is larger than the test errors for the previously discussed interpolant cases (less than $0.69 \%$). Although, given we are interpolating using only three points on a non-linear manifold, our method seems to perform reasonably well.
    
\subsubsection{Extrapolation}
\begin{figure}[htbp]
        \centering
        \begin{overpic}[width=0.8\textwidth]{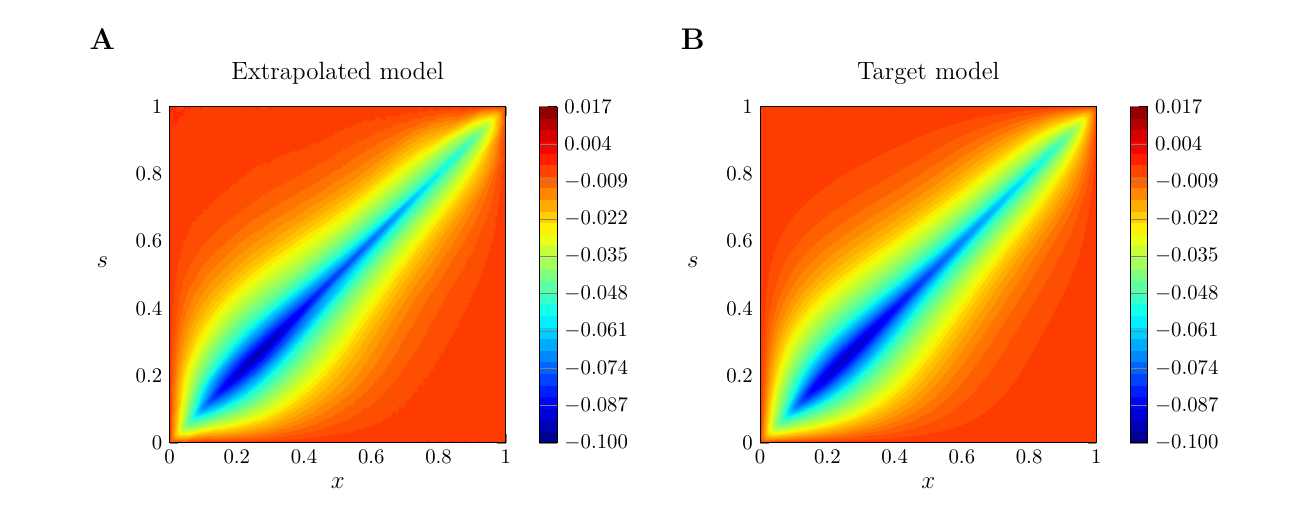}
        \end{overpic}
        \cprotect\caption{(A) Extrapolated Green's function ($\epsilon_{\text{test}} = 2.28 \%$). (B) The target Green's function approximated by \verb|chebgreen| at $\theta_*=9$.}
        \label{airy_extrapolation}
    \end{figure}

    In order to demonstrate the ability of our method to generalize out of the neighborhood of the approximated Green's functions, we compute \verb|chebgreen| approximations at $\theta_1 = 6.0$, $\theta_2 = 7.0$, and $\theta_3 = 8.0$ and extrapolate to $\theta_* = 9.0$. This satisfactory performance is due to our having learned the governing solution operator (i.e., Green's function). A visual comparison between the extrapolated model and a target model constructed with data at $\theta_* = 9.0$ is shown in \cref{airy_extrapolation}. The extrapolated Green's function at $\theta_*$ has a test error of $\epsilon_{\text{test}} = 2.28 \%$.
    
\subsection{Fractional Laplacian}
As a final example, we consider a non-local operator in the form of a one-dimensional fractional Laplacian having periodic boundary conditions:
\begin{align}
    (-\Delta)^\theta u = f, \quad u \left(-\dfrac{\pi}{2}\right) = u\left(-\dfrac{\pi}{2}\right), \quad x \in \left[-\dfrac{\pi}{2},\dfrac{\pi}{2} \right]
    \label{fractional-eq}
\end{align}
where $0 < \theta < 1$ is the fractional order. To generate the dataset for this problem, we use a Fourier spectral collocation to solve \cref{fractional-eq}. The forcing and system responses are sampled at a slightly smaller uniform 1D grid, with $N_x = N_s = 600$ samples. For chebfun, we set $\varepsilon_x^{\text{cheb}} = \varepsilon_s^{\text{cheb}} = 1 \times 10^{-9}$ as the Green's function tends to be non-differentiable throughout the domain which makes it difficult to resolve this to \verb|float64| precision.

\begin{figure}[htbp]
    \centering
    \begin{overpic}[width=\textwidth]{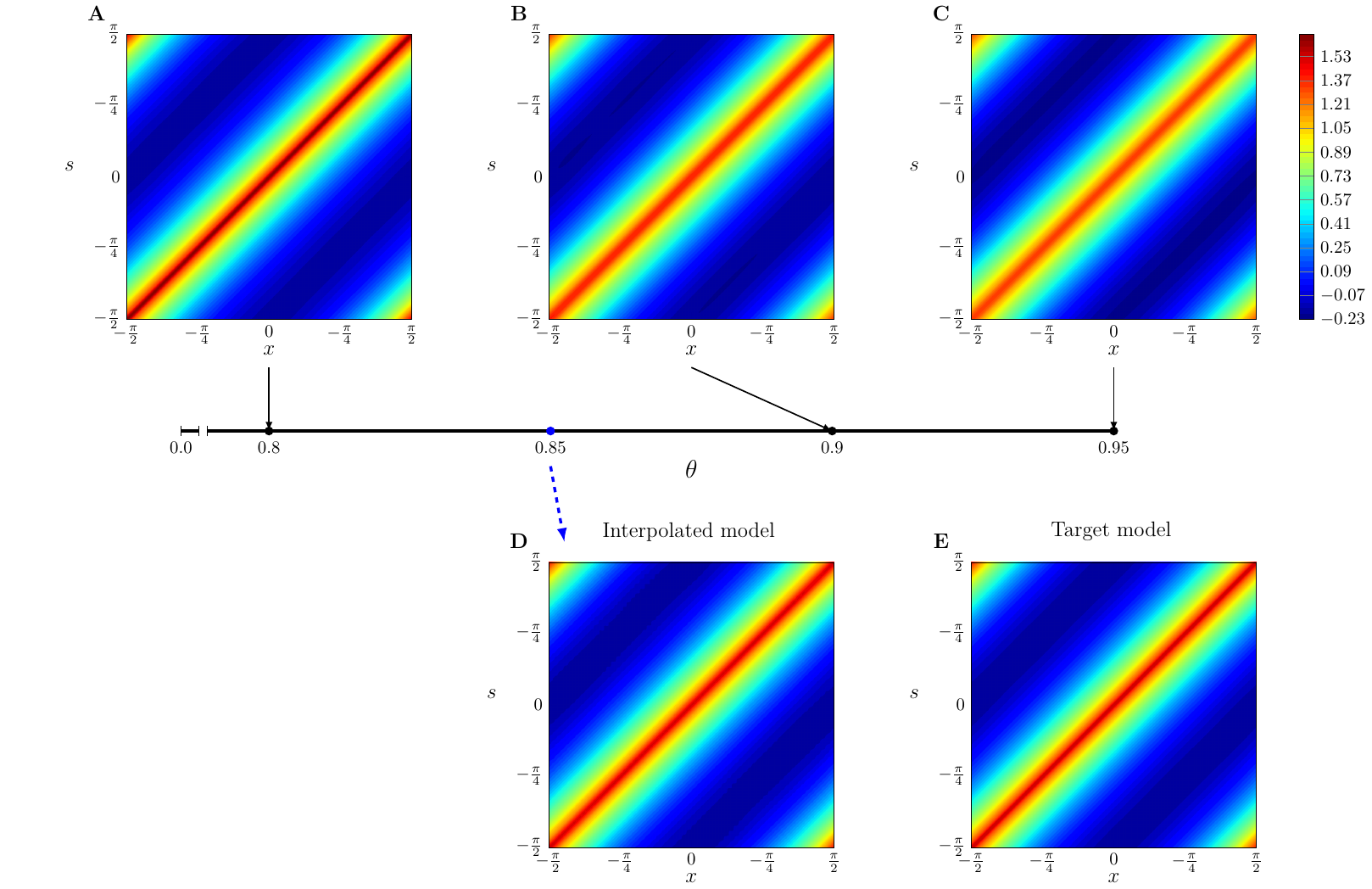}
    \end{overpic}
    \cprotect\caption{(A-C) Approximation of the Green's functions associated with the Fractional Laplacian problem at $\theta=0.8,0.9,0.95$ used by the interpolation scheme. (D) Interpolated Green's function at $\theta_*=0.85$ $\epsilon_{\text{test}} = 0.99 \%$. (E) The target Green's function approximated by \verb|chebgreen|.}
    \label{fractional_laplacian}
\end{figure}

We follow the same procedure of approximating three Green's functions at parameter values $\theta_1 = 0.8$, $\theta_2 = 0.9$, and $\theta_3 = 0.95$ and interpolating to $\theta_* = 0.9$. The test error for the interpolated model is equal to $\epsilon_{\text{test}} = 0.99 \%$ which is comparable to the test error for the interpolant cases (less than $0.95 \%$). A visual comparison of the interpolated model against the target model is shown in \cref{fractional_laplacian}C-D.

\section{Conclusions}
In this work, we propose an extension of the previous work \cite{Praveen2023_EGF} on learning and interpolating Green's function from data. Some notable improvements include the inclusion of \verb|greenlearning| \cite{Boulle2021_Greenlearning} to learn an approximation of Green's function in form of a continuous, bivariate function and the use of our python implementation of \verb|chebfun| library \cite{Chebfun}, \verb|chebpy|, to learn a Singular Value Expansion instead of Singular Value Decomposition, in the previous work \cite{Praveen2023_EGF}. This allows the new method to learn a representation which is not tied to a particular discretization of the domain, making it truly mesh-independent. Finally, the novel generalization of the previous interpolation algorithm for orthonormal matrices on Stiefel manifold to the interpolation of Quasimatrices on the infinite dimensional analogue of Stiefel manifold, provides a way to seamlessly use the machinery we have created in order learn the solution operator for 1D parametric linear partial differential equations. We present multiple examples in order to demonstrate that the method learns high-fidelity approximations to the underlying Green's function of a linear operator with an \textit{order of magnitude} lower data requirement compared to the previous work. We also demonstrate that the proposed method is robust against large amounts of noise in the dataset. This, along with the easy to use library which is available online makes it a useful tool for use in discovering the solution operators in realistic experimental settings.

\cprotect\edit{A primary direction for future investigation is the extension of our method to higher-dimensional problems. This requires careful consideration, as Eckart-Young-Mirsky theorem is not valid for tensors of order greater than two. One strategy would be using the same approach as the \verb|chebfun3| \cite{Hashemi2017_chebfun3} framework: Using the functional Tucker representation \cite{Dolgov2021_TuckerFuncApprox} to create continuous approximations of the Green's functions with a Chebyshev basis, followed by the continuous analogue of the Tensor Train decomposition \cite{Gorodetsky2019_TensorTrainContinuous} to construct a representation we can interpolate with our proposed manifold interpolation method. The other strategy would be to  construct a reduced order model using Deep Learning \cite{Brivio2024_PODDeep}.}

\subsection{Summary of errors for all problems}
The test errors for all the interpolant, interpolated, and target Green's functions are summarized in \cref{tb:summary-errors}. We also compute the test errors when the dataset has $50 \%$ output noise (noise level $\zeta = 0.5$). Note that here we learn the Green's function with the noisy dataset but we test with a clean dataset; to demonstrate that we indeed learn the underlying solution operator. The test error on the noisy dataset for these cases are available in \cref{apx:realstic-measures}.
\label{sec:conclusion}

\begin{table}[htbp]
    \centering
    \begin{tabular}{cccccc}
        \toprule
        Problem                 & $\theta_1$   & $\theta_2$    & $\theta_3$    & Interpolated  & Target \\
        \midrule
        Airy	                & 0.38 (1.87)  & 0.69 (1.91)    & 0.62 (2.02)   & 2.76 (3.28)   & 0.76 (1.84) \\
        Fractional Laplacian	& 0.51 (2.38)  & 0.95 (2.81)   & 0.65 (1.87)   & 0.99 (2.94)   & 0.41 (2.21) \\
        Advection Diffusion	    & 0.24 (1.57)  & 0.26 (1.49)   & 0.29 (1.49)   & 0.54 (1.61)   & 0.3 (1.54) \\
        \bottomrule
    \end{tabular}
    \caption{Summary of the test error for the Airy, Fractional Laplacian, and Advection-Diffusion problems. The error for experiments with noisy dataset are shown in parenthesis.}
    \label{tb:summary-errors}
\end{table}

\section*{Data availability}

\noindent The code used to produce the numerical results is publicly available on

\begin{itemize}
    \item GitHub: \url{https://github.com/hsharsh/chebgreen}
    \item GitLab: \url{https://gitlab.com/praveenharsh/chebgreen}
\end{itemize}

for reproducibility purposes.

\section*{Acknowledgments}
This work was supported by the SciAI Center, and funded by the Office of Naval Research (ONR), under Grant Numbers N00014-23-1-2729 and N00014-22-1-2055.

\bibliographystyle{elsarticle-num}
\bibliography{main}

\newpage
\appendix
\section{Proof for interpolating quasimatrices on tangent space of $H=(L^2(\Omega))^K$ space}

\begin{customthm}{1}\label{theorem-one}
    The map $P_{\Phi}\colon H\rightarrow T_\Phi S_K(H)$ given by $\Psi\mapsto \Psi - \Phi\operatorname{sym}(\Phi^*\Psi)$ is the orthogonal projection onto the tangent space of the Stiefel manifold $S_K(H)$ at $\Phi,$ where $\operatorname{sym}(\Phi^* \Psi) \coloneq \frac{1}{2}(\Phi^* \Psi + \Psi^* \Phi).$
\end{customthm}

\begin{proof}
    Let $A = \Phi^* \Psi$ and $B=\Psi^* \Phi.$ First we show that $P_\Phi$ is a projection operator, i.e. that $P_\Phi \circ P_\Phi = P_\Phi.$ For $\Psi\in H,$ we have
    $$P_\Phi(P_\Phi(\Psi))= \Psi - \frac{1}{2}\Phi(A+B)-\frac{1}{2}\Phi\left([\Psi - \frac{1}{2}\Phi(A+B)]^* \Phi + \Phi^* [\Psi - \frac{1}{2}\Phi(A+B)] \right)$$
    $$= \Psi - \frac{1}{2}\Phi(A+B)-\frac{1}{2}\Phi\left(\Psi^* \Phi -[\frac{1}{2}\Phi(A+B)]^* \Phi + \Phi^* \Psi -\Phi^* [\frac{1}{2}\Phi(A+B)] \right) $$
    $$=\Psi - \frac{1}{2}\Phi(A+B)-\frac{1}{2}\Phi\left(A -\frac{(A+B)}{2}\Phi^* \Phi +B -\Phi^* \Phi \frac{(A+B)}{2} \right)$$
    Since $\Phi$ is an element of the Stiefel manifold, it satisfies $\Phi^* \Phi = I_K.$ So all of the terms in the last line to the right of $\Psi - \frac{1}{2}\Phi(A+B)$ cancel out, and it follows that $P_\Phi  P_\Phi = P_\Phi$ as desired.

    Next, we must show that $P_\Phi$ is self-adjoint in order to establish that it is an orthogonal projection. For any $\Psi, \Gamma\in H,$ we have
    $$\tr{P_\Phi (\Psi)^* \Gamma} = \tr{\Psi^* \Gamma - \frac{1}{2} \left[\Phi (\Phi^* \Psi + \Psi^* \Phi)\right]^* \Gamma},$$
    and
    $$\tr{\Psi^* P_\Phi(\Gamma)} = \tr{\Psi^* \Gamma - \frac{1}{2}\Psi^* \left[\Phi(\Phi^* \Gamma + \Gamma^* \Phi) \right]}.$$
    So it suffices to show that
    \begin{equation}
        \tr{\left[ \Phi(\Phi^* \Psi + \Psi^* \Phi)\right]^* \Gamma} = \tr{\Psi^* \left[\Phi(\Phi^* \Gamma + \Gamma^* \Phi) \right]}.
    \end{equation}
    To see that left side of Eq. (A.1) satisfies
    \begin{equation}
           \operatorname{tr}\left(\left[ \Phi(\Phi^* \Psi + \Psi^* \Phi)\right]
           ^*\Gamma \right) = \operatorname{tr}\left((\Phi^* \Psi +\Psi^* \Phi) (\Phi^* \Gamma)\right) 
    \end{equation}
    consider the left side of Eq. (A.2),
    $$ \operatorname{tr}\left( [\Phi(\Phi^*\Psi + \Psi^*\Phi)]^*\Gamma \right) = \operatorname{tr}\left([\Phi(\Phi^*\Psi)]^*\Gamma+[\Phi(\Psi^*\Phi)]^*\Gamma \right), $$
    note that computing the term in the left of trace argument, $[\Phi(\Phi^*\Psi)]^*\Gamma,$ gives a $k\times k$ matrix $C$ such that
    $$C_{ij} = \left(\sum_{i=1}^k \Phi_i(\Phi_j,\Psi_k)_{L^2(\Omega)},\Gamma_j \right)_{L^2(\Omega)}.$$
    Expanding right side of Eq. (A.2), $$\operatorname{tr}\left((\Phi^* \Psi +\Psi^* \Phi) (\Phi^* \Gamma)\right) = \operatorname{tr}\left((\Phi^* \Psi)(\Phi^*\Gamma) +(\Psi^* \Phi) (\Phi^*\Gamma)\right)$$
    and multiplying out the term 
    $(\Phi^*\Psi)(\Phi^*\Gamma)$, we find that this is equal to $C=[\Phi(\Phi^*\Psi)]^*\Gamma.$ Doing the same computation for the other term, we conclude that Eq. (A.2) holds. 
    
    Similarly, the right side of Eq. (A.1) gives
    $$\tr{\Psi^* \left[\Phi(\Phi^* \Gamma + \Gamma^* \Phi) \right]} = \tr{(\Psi^* \Phi)(\Phi^* \Gamma + \Gamma^* \Phi)}$$
    Using the fact that $(\Xi^* \Omega)^T = \Omega^* \Xi$ for any $\Xi, \Omega\in H,$ which comes from the definition of the outer product on $H$ in Eq. (8), it follows that Eq. (A.1) holds. Hence $P_\Phi$ is an orthogonal projection.

    Finally, to see that the image of $P_\Phi$ is all of $T_\Phi S_K(V),$ consider the operator $P^{\perp}_\Phi$ defined by the mapping $\Psi\mapsto \Phi\operatorname{sym}(\Phi^* \Psi).$ The normal space may be characterized as the set of all elements of $H$ of the form $\Phi M$ where $M$ is a symmetric matrix. Clearly any element of the normal space is the image of some $\Psi$ under this mapping. Since we have $\Psi = P_\Phi(\Psi) + P^{\perp}_\Phi(\Psi)$ for any $\Psi\in H,$ it follows that $P_\Phi$ is the orthogonal projection onto $T_\Phi S_K(H).$ 
\end{proof}
\section{Error analysis}
\begin{customlem}{1}\label{lemma-one}
    The radius of injectivity of the Riemannian exponential map on the Hilbert Stiefel manifold \(M\) of \( H = (L^2(\Omega))^K\) is \(\pi\).
\end{customlem}

\begin{proof}
    From Zimmermann and Stoye, we know that this is true for the Stiefel manifold \(S_K(\mathbb{R}^n)\) \cite{stoye2024injectivityradiusstiefelmanifold}. Furthermore, Harms and Mennucci showed that a minimal geodesic on \(S_K\) is contained in a an isometrically embedded submanifold of \(S_K\) \cite{Harms2012_Manifolds}. By Theorem 3 (5) from Harms and Menucci \cite{Harms2012_Manifolds}, this submanifold happens to be \(S_{K}(V)\): the \(K\)-Stiefel manifold of some \(2K\) dimensional subspace \(V\) of \(H\). Since this geodesic is contained in a finite dimensional Stiefel manifold, it follows that the radius of injectivity of the Hilbert Stiefel manifold \(S_K\) is also \(\pi\).
\end{proof}

\begin{customprop}{1}\label{prop-one}  
    Let \(S_K\) be the Hilbert-Stiefel manifold endowed with the Riemannian metric induced from
    \[
    H = \bigl(L^2(\Omega)\bigr)^K,
    \]
    and let \(\phi \in S_K\). Suppose the neighborhood \(U\) of the origin in \(T_\phi S_K\) contains only tangent vectors \(v\) such that \(\|v\|_H <\pi\). Let \(P_\phi: H \to T_\phi S_K\) be the orthogonal projection onto \(T_\phi S_K\) . Then there exists a constant \(C>0\) such that, for every \(y \in W\) with \(y=\exp_\phi(v)\), where \(W\) is the image of the exponential map on \(U\), we have
\[
\|v-P_\phi(y-\phi)\|_H\le C\,\|y-\phi\|_H^3.
\]
\end{customprop}

\begin{proof}
Let \(y=\exp_\phi(v)\) with \(v\in U\subset T_\phi S_K\). We know that \(U\) is within the radius of injectivity because the metric on \(S_K\) is induced by \(H\), and hence by the Taylor expansion of \(\exp_\phi\) in normal coordinates, which exist in $U$ since $U$ is contained within the radius of injectivity of the exponential map, the first derivative is the identity and the second derivative vanishes \cite{Boumal_2023}. Thus
\[
\exp_\phi(v)= \phi+v+\frac{1}{6}\,D^3\exp_\phi(0)[v,v,v] + r(v),
\]
with a remainder satisfying
\[
\|r(v)\|_H = O(\|v\|_H^4).
\]
Subtracting \(\phi\) from both sides gives
\[
y-\phi = v+\frac{1}{6}\,D^3\exp_\phi(0)[v,v,v] + r(v).
\]
Since \(v\in T_\phi S_K\) and \(P_\phi\) is the orthogonal projection onto \(T_\phi S_K\) (with \(P_\phi(w)=w\) for all \(w\in T_\phi S_K\)), it follows that
\[
P_\phi(y-\phi) = v+ P_\phi\Bigl(\frac{1}{6}\,D^3\exp_\phi(0)[v,v,v] + r(v)\Bigr).
\]
By the non-expansivity of \(P_\phi\), we have
\[
\|P_\phi(y-\phi)-v\|_H \le \frac{1}{6}\,\|D^3\exp_\phi(0)[v,v,v]\|_H + \|r(v)\|_H.
\]
Since \(D^3\exp_\phi(0)\) is a bounded trilinear map on \(U\), there is a constant \(C_3>0\) such that
\[
\|D^3\exp_\phi(0)[v,v,v]\|_H \le C_3\,\|v\|_H^3.
\]
Thus,
\[
\|P_\phi(y-\phi)-v\|_H \le \frac{C_3}{6}\,\|v\|_H^3 + O(\|v\|_H^4).
\]
Finally, because by the local rigidity of \(\exp_\phi\) we have
\[
\|y-\phi\|_H = \|v\|_H + O(\|v\|_H^3),
\]
it follows (after possibly adjusting the constant) that
\[
\|P_\phi(y-\phi)-v\|_H = O(\|v\|_H^3) = O(\|y-\phi\|_H^3).
\]
\end{proof}

\begin{customprop}{2}\label{prop-two}
    Let \(M\subset H\) be an embedded Hilbert manifold in a Hilbert space \(H\), so that each tangent space \(T_{\phi}M\) is a linear subspace of \(H\). Suppose that 
\[
R: TM \to M,\quad (\phi,\eta) \mapsto R(\phi,\eta)
\]
is a \(C^2\) retraction, that is that:
\begin{enumerate}
  \item[(i)] \(R(\phi,0)=\phi\).
  \item[(ii)] \(DR(\phi,0)[\eta]=\eta\) for all \(\eta\in T_{\phi}M\).
  \item[(iii)] \(R(\phi,\cdot)\) is \(C^2\). Hence, by continuity there exists a constant \(L>0\) (depending on \(\epsilon\)) such that
  \[
  \|D^2R(x,0)[\eta,\eta]\|_H \le L\,\|\eta\|_H^2\quad\text{for all } \eta\in B_\epsilon(0).
  \]
\end{enumerate}
Then for every \(x\in M\) and every \(\eta\in T_xM\) with \(\|\eta\|_H<\epsilon\), there exists \(\delta>0\) (with \(\delta\le \epsilon/\|\eta\|_H\)) such that for all \(t\) with \(|t|<\delta\) we have
\[
\|R(x,t\eta) - (x+t\eta)\|_H \le L\,t^2\,\|\eta\|_H^2.
\]
\end{customprop}

\begin{proof}
    Since \(R(\phi,\cdot):T_{\phi}M\to M\) is smooth on \(B_\epsilon(0) \), Taylor's theorem in the Banach space \( T_\{\phi\}M \) (with norm \(\|\cdot\|_H\) implies that for each \(\phi\in M\) and each \(\eta\in T_{\phi}M \) with \(\|\eta\|_H<\epsilon\) there exists a remainder term \(r(t\eta)\) with
\[
R(\phi,t\eta) = R(\phi,0) + t\,DR(\phi,0)[\eta] + \frac{t^2}{2}\,D^2R(\phi,0)[\eta,\eta] + r(t\eta),
\]
and 
\[
\|r(t\eta)\|_H = O\bigl(t^3\|\eta\|_H^3\bigr)\quad\text{as }t\to 0.
\]
By (i) and (ii), we have
\[
R(\phi,0)=\phi,\quad DR(\phi,0)[\eta] = \eta.
\]
Thus,
\[
R(\phi,t\eta) = \phi + t\eta + \frac{t^2}{2}\,D^2R(\phi,0)[\eta,\eta] + r(t\eta).
\]
Subtracting \(\phi+t\eta\) and taking the \(H\)-norm, we obtain
\[
\|R(\phi,t\eta) - (\phi+t\eta)\|_H \le \frac{t^2}{2}\,\|D^2R(\phi,0)[\eta,\eta]\|_H + \|r(t\eta)\|_H.
\]
By assumption (iii),
\[
\|D^2R(\phi,0)[\eta,\eta]\|_H \le L\,\|\eta\|_H^2.
\]
Hence,
\[
\|R(\phi,t\eta) - (\phi+t\eta)\|_H \le \frac{L}{2}\,t^2\,\|\eta\|_H^2 + \|r(t\eta)\|_H.
\]
Since \(\|r(t\eta)\|_H = o(t^2\|\eta\|_H^2)\), for all \(\varepsilon_0>0\) there exists ,\(\delta_0>0\) such that if \(0<|t|<\delta_0\), then
\[\|r(t\eta)\|_H \le \varepsilon_0\,t^2\|\eta\|_H^2.
\]
Taking \(\varepsilon_0 = L/2\) yields
\[
\|r(t\eta)\|_H \le \frac{L}{2}\,t^2\|\eta\|_H^2
\quad\text{whenever }|t|<\delta_0.
\]

Thus, for \(|t|<\delta\),
\[
\|R(\phi,t\eta) - (\phi+t\eta)\|_H \le \frac{L}{2}\,t^2\,\|\eta\|_H^2 + \frac{L}{2}\,t^2\,\|\eta\|_H^2 = L\,t^2\,\|\eta\|_H^2.
\]
This completes the proof.
\end{proof}

\begin{customlem}{2}\label{lemma-two}
    Let
\[
H = \left( L^2(\Omega) \right)^K
\]
be a Hilbert space and let $S_K$
be the Hilbert Stiefel manifold of $H$. Define the map
\[
qf: U\subset H \to S_K,
\]
by applying the Gram--Schmidt process, that is, define
\[
q_1(\phi_1) = \frac{\phi_1}{\|\phi_1\|_H},
\]
and for \(j=2,\dots,K\), recursively set
\[
p_j(\phi_1, \dots, \phi_j) = \phi_j - \sum_{i=1}^{j-1} \langle \phi_j, q_i(\phi_1,\dots,\phi_i) \rangle_H\, q_i(\phi_1,\dots,\phi_i),
\]
\[
q_j(\phi_1,\dots,\phi_j) = \frac{p_j(\phi_1,\dots,\phi_j)}{\|p_j(\phi_1,\dots,\phi_j)\|_H}.
\]
Then, the mapping
\[
qf(\phi) = \bigl( q_1(\phi_1), \; q_2(\phi_1,\phi_2), \; \dots, \; q_K(\phi_1,\dots,\phi_K) \bigr)
\]
is smooth on the open set
\[
U = \{ \phi \in H : \|\phi_i\|_H > 0 \text{ for } i=1,\dots, K, \text{ and } \{\phi_1,\dots,\phi_K\} \text{ is linearly independent} \}.
\]
\end{customlem}

\begin{proof}
We view the process as a composition of maps:
First, the normalization \(q_1(\phi_1) = \phi_1/\|\phi_1\|_H\) is smooth on the set where \(\|\phi_1\|_H > 0\).
For each \(j \ge 2\), given that \(q_1,\dots,q_{j-1}\) are smooth, the map 
    \[
    \phi_j \mapsto \operatorname{proj}_{\operatorname{span}\{\phi_1,\dots,\phi_{j-1}\}}(\phi_j) = \sum_{i=1}^{j-1} \langle \phi_j, q_i \rangle_H\, q_i,
    \]
    is smooth since the inner product and scalar multiplication are smooth operations in a Hilbert space.
    Then, subtracting this projection from \(\phi_j\) is linear and bounded and hence smooth; the normalization step,
    \[
    q_j = \frac{p_j}{\|p_j\|_H},
    \]
    is smooth on the set of quasimatrices satisfying \(p_j\neq 0\), which holds under the assumption that the input has full rank.

Since each \(q_j\) is smooth and the overall mapping \(qf\) is given by the composition of these maps, \(qf\) is smooth on \(U\).
\end{proof}

\section{Test error for interpolation cases with noisy dataset}

\begin{table}[htbp]
\centering
\begin{tabular}{cccccc}
\toprule
Problem                 & $\theta_1$ & $\theta_2$ & $\theta_3$ & Interpolated & Target \\
\midrule
Airy	                & 6.74       & 6.76       & 6.83       & 7.46         & 6.78 \\
Fractional Laplacian    & 7.52       & 7.55       & 7.48       & 7.58         & 7.49 \\
Advection Diffusion     & 6.63       & 6.61       & 6.69       & 6.65         & 6.66 \\
\bottomrule
\end{tabular}
\caption{Summary of the empirical error for the Airy, Fractional Laplacian, and Advection-Diffusion problems, in the case where the models are learned and tested on a noisy dataset.}
\label{apx:realstic-measures}
\end{table}

\end{document}